\pdfoutput=1
\PassOptionsToPackage{table}{xcolor}
\PassOptionsToPackage{dvipsnames}{xcolor}

\documentclass[11pt]{article}

\usepackage[final]{acl}
\usepackage{xcolor}
\usepackage[normalem]{ulem}

\usepackage{times}
\usepackage{latexsym}
\usepackage{pgfplots}
\usepgfplotslibrary{groupplots}

\usepackage{lipsum}
\usepackage{array}
\usepackage{multicol}
\usepackage{longtable}

\usepackage{comment}

\usepackage[T1]{fontenc}

\usepackage{multirow}

\usepackage[utf8]{inputenc}

\usepackage{microtype}

\usepackage{inconsolata}

\usepackage{graphicx}
\usepackage{linguex}
\usepackage{booktabs}
\usepackage{xcolor}

\title{\textit{They want to pretend not to understand:} The Limits of Current LLMs in Interpreting Implicit Content of Political Discourse}

\author{Walter Paci \\
  University of Florence \\
  \texttt{walter.paci@unifi.it} \\\And
  Alessandro Panunzi \\
  University of Florence \\
  \texttt{alessandro.panunzi@unifi.it} \\\And
  Sandro Pezzelle\\
  University of Amsterdam \\
  \texttt{s.pezzelle@uva.nl} \\}

\begin{document}
\maketitle
\begin{abstract}
Implicit content plays a crucial role in political discourse, where speakers systematically employ pragmatic strategies such as implicatures and presuppositions to influence 
their audiences. 
Large Language Models (LLMs) have demonstrated strong performance in tasks requiring complex semantic and pragmatic understanding, highlighting their potential for detecting and explaining the meaning of implicit content.
However, their ability to do this within political discourse remains largely underexplored. Leveraging, for the first time, the large IMPAQTS corpus, which comprises Italian political speeches with the annotation of manipulative implicit content, we propose methods to test the effectiveness of LLMs in this challenging problem. Through a multiple-choice task and an open-ended generation task, we demonstrate that all tested models struggle to interpret presuppositions and implicatures. 
We conclude that current LLMs lack the key pragmatic capabilities necessary for accurately interpreting highly implicit language, such as that found in political discourse. At the same time, we highlight promising trends and future directions for enhancing model performance. We release our data and code at: \url{https://github.com/WalterPaci/IMPAQTS-PID}
\end{abstract}

\section{Introduction} \label{Intro}
Implicit and vague language is 
pervasive in everyday life communication. Consider the language used in political contexts: politicians often use implicit or vague content, relying on the listener's assumptions, background knowledge, and shared contextual knowledge to convey meaning indirectly. This is evident in the use of vague terms, allusions, and rhetorical devices that allow for multiple interpretations. Such language creates an environment where meanings can be conveyed subtly, often with the intent of persuading, manipulating, or reinforcing certain ideologies~\cite{morency2008explicitness}.
Through the careful use of implicit language, politicians can appeal to the audience’s emotions, assumptions, and cultural references, all while avoiding explicit commitments that could later be challenged or disproved \cite{vallauri2019lingua}.
Consider the following example (also featured in the title of this work): \emph{They want to pretend not to understand.}\footnote{This example is the English translation of the Italian sentence \textit{Vogliono far finta di non capire}, that is part of the data used in this work. In the main paper, for simplicity, we only provide English translations. The original Italian texts are available in Appendix~\ref{sec:examples}, \ref{context survey} and \ref{sec:prompt_templates}. In this section, examples may sometimes be adapted for illustrative purposes.} 
By saying this, the speaker implies that the people being addressed pretend they do not understand something obvious, while, in reality, they understand it perfectly well; moreover, the speaker is also subtlety hinting that they are pretending so \textit{for a reason}, guiding and manipulating the audience to reach a certain reaction or effect. This pragmatic phenomenon, where the speaker says one thing to imply another, is called an implicature.
Implicatures are among the most well-studied phenomena of implicit language~\cite{grice19901975,sperber1987precis}, alongside presuppositions~\cite{strawson1964identifying,garner1971presupposition,ducrot1972dire}. As both are pervasive in political discourse, which makes it an ideal domain for analyzing implicit communication strategies~\cite{van1992discourse, vallauri2020comunicazione}.

The advent of Large Language Models (LLMs) has revolutionized NLP, bringing unprecedented progress to tasks involving semantic understanding~\cite{wang-etal-2018-glue,williams-etal-2018-broad}, question answering~\cite{mihaylov-etal-2018-suit}, and pragmatic interpretation~\cite{zheng2021grice,sravanthi-etal-2024-pub,hu-etal-2023-fine, kim2023pope,ruis2024goldilocks}. 
Overall, LLMs have been shown to perform excellently on semantic tasks and to exhibit potential in pragmatic tasks, particularly when they involve artificially constructed stimuli.
However, whether they can handle tasks that require making pragmatic inferences on naturalistic discourse data, such as political discourse, remains an open question.
In this work, we specifically address this research question and test whether current LLMs can explain the meaning of manipulative implicit content in political discourse, such as the implicature in the example above. Leveraging, for the first time in the context of NLP research, the large IMPAQTS corpus~\cite{cominetti-etal-2024-impaqts}, which includes transcribed Italian political speeches with expert annotations of various types of implicit content, we propose methods to assess the pragmatic effectiveness of several LLMs, state-of-the-art for the Italian language.
Our contributions are as follows:

\begin{itemize}
    \item We propose \textbf{a challenging task for LLMs}, i.e., to clarify, through an explanation, the meaning of a manipulative implicit content found in political discourse; e.g., that in the passage \emph{They want to pretend not to understand}, the speaker implies that these people understand perfectly well. 
    \item We release \textbf{a novel dataset}, based on the IMPAQTS corpus~\cite{cominetti-etal-2024-impaqts}, containing 30K implicit passages (implicatures and presuppositions) with expert-based explanations, as well as the surrounding linguistic context necessary to correctly interpret the implicit content (empirically validated through human expert annotation). 
    \item Through a multiple-choice task and an open-ended generation task, we demonstrate that \textbf{all tested models struggle} to interpret manipulative presuppositions and implicatures. In the multiple-choice setup, the best-performing model falls more than 20 accuracy points short of the estimated ceiling; in the open-ended generation setup, it provides a fully correct explanation in only one-fourth of the cases.
    \item Despite largely unsatisfactory results, we show that using \textbf{Chain-of-Thought (CoT) improves performance} in the open-ended generation setup, suggesting that reasoning mechanisms can aid in solving complex pragmatic tasks. We also propose other directions, such as \textbf{embedding external knowledge} about the politician and their political affiliation to further enhance model performance.
\end{itemize}

\section{Related Work}\label{related_work}
\subsection{Implicit Language and LLMs}

As LLMs power conversational agents that interact with human users, they must be capable of understanding implicit content.
Prior research has explored LLMs' capacity to interpret non-literal meanings in linguistic discourse.
\citet{jeretic-etal-2020-natural} investigated the emergence of pragmatic understanding by analyzing how pre-trained language models compute scalar implicatures and presuppositions. They found strong evidence that 
models such as BERT learned scalar implicatures for quantifiers \textit{some} and \textit{all}, but struggled with other scalar pairs, treating them as synonymous. For presuppositions, models failed to recognize some presupposition triggers, e.g., the verb \textit{to stop} when it presupposes an action that used to be made, as in \textit{John stopped smoking}. \citet{zheng2021grice} proposed the first evaluation of language models on five types of implicatures derived from~\citeauthor{grice19901975}'s~\citeyearpar{grice19901975} conversational maxims, finding that models performed relatively well on a multiple-choice task, where they had to choose which of the given options explained the implicature in a dialogue, but struggled significantly with conversational reasoning, suggesting they do not fully understand conversational context.
On a similar track,
\citet{hu-etal-2023-fine} evaluated pre-trained LMs without task-specific fine-tuning, incorporating Gricean implicatures as one of their tested phenomena among many others. Their results suggested that language models can infer pragmatic meanings, though they left unresolved whether this is due to linguistic cues or cognitive processes, highlighting the need for further research on their connection in pragmatic reasoning.
\citet{kim2023pope} evaluated whether LLMs can understand conversational implicatures by prompting them to provide binary answers to specific scenarios. They demonstrated that while LLMs exhibited some implicit understanding, their performance improved significantly when guided through the reasoning process
using chain-of-thought prompting.
More recently, \citet{ruis2024goldilocks} designed a protocol to evaluate LLMs on binary implicature resolution, highlighting a significant gap between humans 
and LLMs.

All these studies investigated plausible naturalistic use of pragmatic language. Nonetheless, they used artificially constructed sentences instead of real examples of pragmatic phenomena found in actual \textit{corpora}.
In our work, we make a significant step forward and use excerpts of semi-spontaneous, ecological political discourse that reflect the complexities of real-world communication. 
Unlike artificially constructed sentences, political discourse is known for capturing the complex interplay of rhetoric, social context, and speaker intent. 

\subsection{Political Language in NLP}

Political language has been the focus of NLP research leveraging large datasets from social media and political speeches to study discourse framing, bias detection, and polarization.

\citet{katre2019nlp} employed computational methods for text analytics and visualization in political speech transcripts. The author used NLP techniques to analyze political speeches and generate graphical visualizations for, among others, word use, lexical dispersion, etc.
In~\citet{huguet-cabot-etal-2020-pragmatics}, NLP methods were used to model political discourse by focusing on metaphor, emotion, and political rhetoric. The authors presented the first joint models that integrate metaphor and emotion detection as auxiliary tasks to enhance the performance of predicting the political perspective of news articles, party affiliation of politicians, and framing of policy issues.  \citet{nemeth2023scoping} presented a methodological review on how researchers have used NLP techniques to study language polarization between 2010 and 2021. The author identified data sources and computational approaches and reviewed the different conceptualizations and operationalizations of polarization.

Recent efforts leveraged LLMs to enhance political discourse understanding.
\citet{Marino_Giglietto_2024}
used LLMs for political discourse annotation, reporting significant improvement on previous methods like topic modeling. Here, LLMs were used to analyze Facebook links related to the 2018 and 2022 Italian elections and perform tasks such as classification and clustering of political content and generation of descriptive labels for these clusters.
\citet{li2024political} proposed Political-LLM, a comprehensive framework for integrating LLMs into political science. This offered unprecedented capabilities for text analysis. At the same time, the authors highlighted several issues related to biases, ethics, and explainability to be considered when using LLMs in political applications.
Despite these advancements, little attention has been paid to automatically explaining implicit content. To our knowledge, we are the first to tackle this problem.

\section{Data}\label{corpus}

\subsection{The IMPAQTS Corpus}

We use, for the very first time in the context of NLP research, data from IMPAQTS \cite{cominetti-etal-2024-impaqts},\footnote{\url{https://impaqts.dilef.unifi.it/}} a corpus of Italian political discourse \cite[in the sense of "discourse by politicians"; see][]{van1997political}, consisting of 1,500 speeches uttered in the Italian language by 150 prominent politicians between 1946 and 2023.
IMPAQTS is a multimodal corpus containing over 800 speeches in video format and around 600 in audio format. The manual transcriptions of these speeches are also available, building up to roughly 2.65 million tokens. In our work, we only use these transcriptions.

A key feature of IMPAQTS, crucial for our work, is the annotation of all the passages that contain \textit{implicit content with some manipulative meaning}.
In more technical terminology, this kind of implicit content is referred to as non-\textit{bona fide} true; these are implicit questionable contents that are not conveyed in good faith but are still non-explicitly understood as true within a given context.  
IMPAQTS defines four types of implicit content: \textit{implicature}, \textit{presupposition}, \textit{topicalization}, and \textit{vagueness}. For an extensive discussion on this theoretical framework, see \citet{lombardi2016exaptation}. Each 
sentence tagged as containing some implicit comment is accompanied by a \textit{comment} written by the IMPAQTS corpus annotators explaining the meaning of the implicit content.\footnote{In the IMPAQTS corpus, each sentence was tagged by 3 independent annotators. These annotations were then validated by a fourth expert who produced a final comment.} Annotators' comments are all in the same, fixed format; e.g., for implicatures, they start with \textit{it implies that\dots}; for presuppositions, \textit{it presupposes that\dots}; for topicalizations, \textit{it considers active in the discourse that\dots}; for vagueness, \textit{it leaves vague that\dots}.

Below, we report a single sample including the comment by the IMPAQTS annotators:

\ex.\textbf{Text with implicit content:} Italy doesn’t need another government led by Monti. The last thing Italy needs is another government that is a slave to the banks!\\\textbf{Comment:} It implies that Monti's government is a slave to the banks.

In this work, we focus on two of the four categories previously presented, i.e., \textit{implicatures} and \textit{presuppositions}. 
We chose not to include topicalization as its understanding and interpretation can heavily depend on the information carried out by its prosodic contour in the spoken language~\cite{frascarelli2008types}, which we cannot leverage when using transcriptions. On the other hand, vagueness was excluded due to its wider, less focused definition.

IMPAQTS only includes monologues, and they belong to six types:
\textit{Assembly speech}, \textit{Rally speech}, \textit{Party assembly speech}, \textit{Statement in presence}, \textit{Broadcast statement}, \textit{New media statement}, and \textit{Operational meeting Interview}. Here, we chose to focus on sentences from Parliamentary and Rally speeches only, in which implicit contents usually do not refer to external contexts, and they can be detected in the speech transcription.
IMPAQTS corpus includes speeches from three time periods: from 1946 to 1972, from 1973 to 1993, and from 1994 to 2022. We experiment with sentences from each of these periods.

\subsection{How Much Linguistic Context Is Needed?}\label{context}

To empirically answer this question, we conducted a human validation study asking 9 expert linguist annotators to assess how many \textit{left-hand} context sentences, i.e., preceding the target sentence, were needed to understand its implicit content. We selected 126 sentences, balanced for both the type of implicit content (63 implicatures and 63 presuppositions) and period (42 sentences for the time range). We then created 3 surveys, each containing 42 samples embedding implicit content.
In each survey, 3 experts were asked to read both the sentence containing the implicit content and its explanation given by the IMPAQTS' annotators. 

Our experts had to assess whether the target sentence provided enough information to understand the explanation in the comment, or if additional preceding linguistic context was needed. If the latter was the case, one extra sentence preceding the target sentence would appear, and the same question would be asked again. By design, our annotators could see up to five left-hand sentences. 
The annotation was carried out on LimeSurvey \cite{LimeSurvey} and allowed to identify, for each sentence, the necessary amount of context for a correct implicit interpretation. The survey details and the inter-annotator agreement are available in Appendix~\ref{context survey}.

\subsection{Experimental Data}

Based on the results of the annotations presented above, we proceeded to construct a dataset to be used in our experiments with LLMs. We consider all the samples in IMPAQTS containing either an implicature or a presupposition. For each of these samples, we retrieved the preceding 4 sentences in the speech. The resulting dataset, which we name IMPAQTS-PID---where \textit{PID} stands for Preupposition and Implicature Dataset---includes 31,822 samples paired with the explanation of the implicit content from IMPAQTS; in particular, 14,932 samples embed an implicature and 16,890 a presupposition.
In Appendix~\ref{sec:descriptive-IMPAQTS-PID} we report some descriptive statistics of our dataset.

\section{Methods} \label{methods}

Using the IMPAQTS-PID dataset presented above, we challenge models to understand a sentence's implicit content. In this section, we describe the models we tested in our work, the experiments we performed to assess their abilities, and the experimental details common to both experiments.

\subsection{Models}\label{models}

We experiment with four models pre-trained with multilingual data, which are therefore suitable to process Italian language: GPT4o-mini \cite{openai2024gpt4omini}, Aya Expanse 8B \cite{cohere2024aya}, LLAMA3.1 8B \cite{meta2024llama31}, and LLAMA3.2 3B \cite{meta2024llama32}. The first model is proprietary, the other three are open-weight.
All these models are relatively small-scale, which has advantages in terms of computational efficiency, costs, and deployment feasibility. Thus, these models strike a good balance by requiring fewer resources, which reduces both hardware costs and energy consumption, while enabling faster inference. We test models' understanding of implicit content through two experiments, which we describe below.
\subsection{Multiple-Choice Generation (MCG)} We perform an experiment where models are presented with four possible explanations IMPAQTS comments (hence, \textit{explanations}) for a given implicit content and must pick the correct one. That is, we frame the problem as a multiple-choice generation (MCG) task. We feed the model with a brief instruction and the four candidates (A, B, C, and D) to choose from, only one of which is correct.
To select challenging \textit{distractors}, we use a method based on topic similarity. We first run a topic modeling analysis on all the explanations in IMPAQTS-PID. We identify 450 topics and a residual category including explanations that do not share a common topic. We then sample the distractors from the same topic class of the target explanation, being it one of the 450 topic-based categories or the residual one. This way, we ensure that the three distractors are incorrect yet plausible, compelling the models to reason over the input rather than relying on shortcuts.
During inference, the order of the four candidates is randomized to avoid biases; that is, the position of the correct answer is roughly balanced across positions A-D.
Given that each instance in our dataset includes a ground-truth answer, we evaluate model performance using plain accuracy.

\subsection{Open-Ended Generation (OEG)}\label{sec:OEG} Framing the problem as an MCG task has various advantages, including being easy to set up and evaluate via accuracy. At the same time, it bears relevant limitations due to the selection of the distractors and the non-naturalistic setting of choosing from available options. In this experiment, we circumvent these limitations and directly query models to generate an explanation for a given implicit content; i.e., we frame the problem as an open-ended generation (OEG) task. We experiment with  three prompting settings, which we briefly describe below:

\paragraph{Zero-shot} The model is queried without any prior examples or detailed guidance and solely relies on its pre-training knowledge. This is the template used for this prompting technique:\\

\noindent \textit{Explain the implicit content of the following text. Consider that it appears in its right-most sentence.} \\\textit{Text:} [sentence embedding implicit content]\\\textit{Implicit Content: \_\_\_\_}

\paragraph{Few-Shot} The prompt includes four illustrative examples, i.e., two implicatures and two presuppositions, including both the text embedding the implicit content and its explanation. As we aim at steering model focus towards manipulative implicit content, we only include "positive" examples (i.e., where an implicature or presupposition is present). We do not include "negative" examples to avoid models focusing on unrelated or unwanted linguistic phenomena.
This is the template used for this prompting technique (for brevity, we only report two \textit{shots} instead of four):\\

\noindent\textit{Text:} [First example text]\\
    \textit{Implicit Content:} [Explanation of the implicit content in the first example text]\\
    \dots\\
    \textit{Text:} [Fourth example text]\\
    \textit{Implicit Content:} [Explanation of the implicit content in the fourth example text]\\
    \textit{Text:} [sentence embedding implicit content]\\
    \textit{Implicit Content: \_\_\_\_}

\noindent\paragraph{Chain-of-Thought (CoT)} The prompt presents an extensive step-by-step explanation outlining the reasoning process necessary to decode the implicit content. We adapt these steps from the detailed instructions given to the IMPAQTS corpus annotators. This is the template used for this prompting technique (shortened for space reasons):\\

\noindent\textit{Implicatures arise in communication whenever the speaker "challenges" one of the four conversational maxims derived from the well-known Cooperative Principle} \dots\\
\textit{We define "presupposition" as any content that is taken for granted by the participants in communication and, more specifically, content conveyed as part of the knowledge already shared by the interlocutor.} \dots\\
\textit{Consider what has just been said about implicatures and presuppositions, and explain, proceeding step by step, what the implicit content in the following text is. Consider that it appears in its right-most sentence.}\\
    \\
    \textit{Text:} [sentence embedding implicit content]\\
    \textit{Implicit Content: \_\_\_\_}

\paragraph{Evaluation} In contrast to the MCG task, where the presence of a ground-truth answer allows for an evaluation based on plain accuracy, assessing model performance here is more complex.
Evaluation based on automatic similarity measures or LLMs-as-judges are potential solutions. However, the task set up introduces uncertainty regarding their reliability.
Given these concerns, we opt for a small-scale but more robust assessment based on human expert evaluation. We devise an evaluation protocol where expert linguists assess the texts containing implicit content, the comment about the implicit content from IMPAQTS, and the implicit content explanation generated by the model. See Appendix~\ref{OEG-eval-instruct} for an example.
The experts are asked to judge the quality of the explanation generated by a model by choosing one between 5 possible options:
\begin{itemize}
    \item \textbf{Totally correct:} the explanation is short and focused on the same topic highlighted in the original comment from IMPAQTS.
    \item \textbf{Correct among various options:} the answer lists many possible explanations; among those, the correct one is mentioned (the correctness of other explanations is not relevant).
    \item \textbf{Partially correct:} the model outputs a short answer that does not fully address the point but captures some elements of the correct explanation.
    \item \textbf{Totally wrong:} a short or long answer that is wrong, i.e., is not about the topic highlighted by the comment from IMPAQTS.
    \item \textbf{Answer not given:} the model refuses to answer, avoids the question or just highlights the sentence containing the implicit content without giving an explanation.
\end{itemize}

\subsection{Experimental Details}
Details on prompting and decoding settings for both tasks, including token limits, temperature parameters, GPUs, and computational time are provided in Appendix \ref{sec:prompt_templates} and \ref{sec:exp_details}.

\section{Results}\label{results}
\subsection{MCG Results} \label{MCG results}
\begin{figure}
    \begin{tikzpicture}
        \begin{axis}[
            width  = 7.5cm,
            height = 6.8cm,
            major x tick style = transparent,
            ybar,
            bar width=20pt,
            ymajorgrids = false,
            ylabel = {Accuracy},
            symbolic x coords={Models},
            xtick = data,
            ytick={0.2,0.3,...,1},
            ymin=0.2,
            ymax=1.0,
            legend style={at={(0.5,-0.12)},
            anchor=north,
            font=\small,
            legend columns=2
            },
            nodes near coords,
            every node near coord/.append style={font=\small},
        ]
        \addplot[bar shift=-2cm][fill=RawSienna!50] coordinates {(Models, 0.6979)};
        \addplot[bar shift=-0.8cm][fill=Orange!50] coordinates {(Models, 0.6202)};
        \addplot[bar shift=0.5cm][fill=TealBlue!50] coordinates {(Models, 0.5618)};
        \addplot[bar shift=1.8cm][fill=RoyalBlue!50] coordinates {(Models, 0.4291)};
    
        \legend{GPT4o-mini, Aya Expanse 8B, LLAMA3.1 8B, LLAMA3.2 3B}
        \addlegendimage{line legend,red,sharp plot}
        \addlegendentry{Chance guessing}
        \draw [line width=1.2pt][color=red!70](rel axis cs: 0,0.065) -- (rel axis cs: 1, 0.065)  node[pos=0.93, above, font=\small] {0.25};
        \addlegendimage{line legend, black, sharp plot, thick}
        \addlegendentry{Estimated ceiling}
        \draw[line width=1.2pt] (rel axis cs: 0,0.90) -- (rel axis cs: 1, 0.90)  node[pos=0.93, above, font=\small] {0.91};
        \addlegendimage{line legend, black, sharp plot, dotted}
        \addlegendentry{n-gram baseline}
        \draw[line width=1.2pt, dotted] (rel axis cs: 0,0.55) -- (rel axis cs: 1, 0.55) node[pos=0.93, above, font=\small] {0.64};
        \end{axis}
    \end{tikzpicture}
    \caption{MCG task. All models perform better than chance level (red line). Only GPT4o-mini consistently surpasses the similarity baseline (dotted line).}
    \label{fig:MCG-Accuracy}
\end{figure}
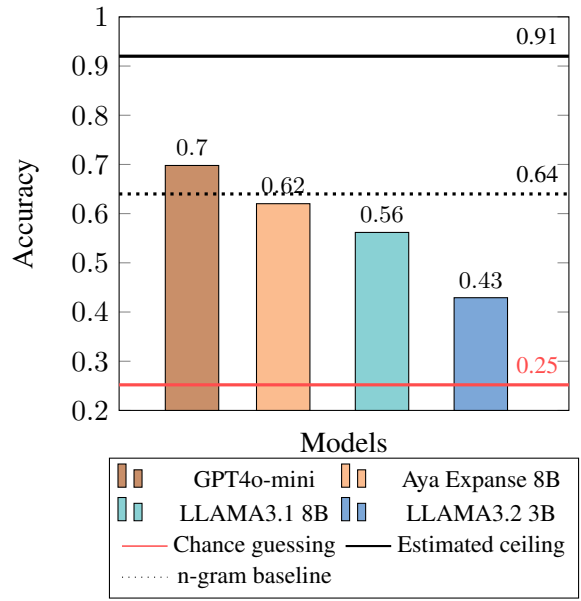

\begin{figure}
    \begin{tikzpicture}
        \begin{axis}[
            width  = 7.5cm,
            height = 6.8cm,
            major x tick style = transparent,
            ybar,
            bar width=20pt,
            ymajorgrids = false,
            ylabel = {Accuracy},
            symbolic x coords={Models},
            xtick = data,
            ytick={0.2,0.3,...,1},
            ymin=0.2,
            ymax=1.0,
            legend style={at={(0.5,-0.12)},
            anchor=north,
            font=\small,
            legend columns=2
            },
            nodes near coords,
            every node near coord/.append style={font=\small},
        ]
        \addplot[bar shift=-2cm][fill=RawSienna!50] coordinates {(Models, 0.654977444)};
        \addplot[bar shift=-0.8cm][fill=Orange!50] coordinates {(Models, 0.579157701)};
        \addplot[bar shift=0.5cm][fill=TealBlue!50] coordinates {(Models, 0.523419058)};
        \addplot[bar shift=1.8cm][fill=RoyalBlue!50] coordinates {(Models, 0.40464420)};
    
        \legend{GPT4o-mini, Aya Expanse 8B, LLAMA3.1 8B, LLAMA3.2 3B}
        \addlegendimage{line legend,red,sharp plot}
        \addlegendentry{Chance guessing}
        \draw [line width=1.2pt][color=red!70](rel axis cs: 0,0.065) -- (rel axis cs: 1, 0.065)  node[pos=0.93, above, font=\small] {0.25};
        \addlegendimage{line legend, black, sharp plot, thick}
        \addlegendentry{Estimated ceiling}
        \draw[line width=1.2pt] (rel axis cs: 0,0.90) -- (rel axis cs: 1, 0.90)  node[pos=0.93, above, font=\small] {0.91};
        \addlegendimage{line legend, black, sharp plot, dotted}
        \addlegendentry{n-gram baseline}
        \draw[line width=1.2pt, dotted] (rel axis cs: 0,0.55) -- (rel axis cs: 1, 0.55) node[pos=0.93, above, font=\small] {0.64};
        \end{axis}
    \end{tikzpicture}
     \caption{\textbf{Hard-negatives setting.} Disaggragated accuracy scores for the MCG Task on the subset of texts with a common topic.}    \label{fig:topic-Accuracy}
\end{figure}
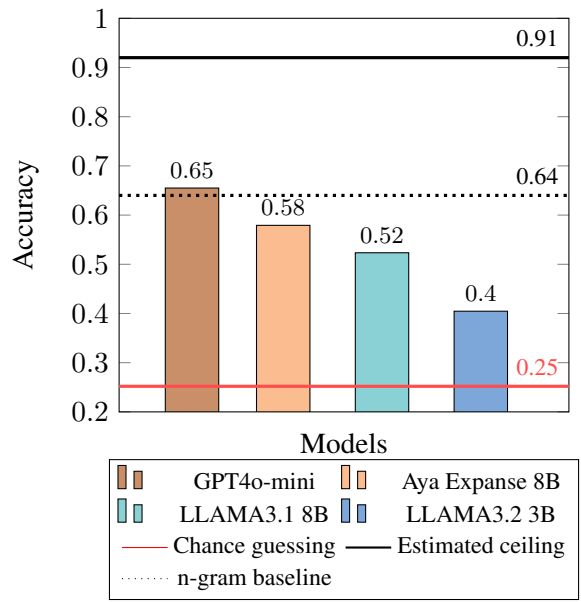 

\begin{figure}
    \begin{tikzpicture}
        \begin{axis}[
            width  = 7.5cm,
            height = 6.8cm,
            major x tick style = transparent,
            ybar,
            bar width=20pt,
            ymajorgrids = false,
            ylabel = {Accuracy},
            symbolic x coords={Models},
            xtick = data,
            ytick={0.2,0.3,...,1},
            ymin=0.2,
            ymax=1.0,
            legend style={at={(0.5,-0.12)},
            anchor=north,
            font=\small,
            legend columns=2
            },
            nodes near coords,
            every node near coord/.append style={font=\small},
        ]
        \addplot[bar shift=-2cm][fill=RawSienna!50] coordinates {(Models, 0.73)};
        \addplot[bar shift=-0.8cm][fill=Orange!50] coordinates {(Models, 0.65)};
        \addplot[bar shift=0.5cm][fill=TealBlue!50] coordinates {(Models, 0.59)};
        \addplot[bar shift=1.8cm][fill=RoyalBlue!50] coordinates {(Models, 0.48)};
    
        \legend{GPT4o-mini, Aya Expanse 8B, LLAMA3.1 8B, LLAMA3.2 3B}
        \addlegendimage{line legend,red,sharp plot}
        \addlegendentry{Chance guessing}
        \draw [line width=1.2pt][color=red!70](rel axis cs: 0,0.065) -- (rel axis cs: 1, 0.065)  node[pos=0.93, above, font=\small] {0.25};
        \addlegendimage{line legend, black, sharp plot, thick}
        \addlegendentry{Estimated ceiling}
        \draw[line width=1.2pt] (rel axis cs: 0,0.90) -- (rel axis cs: 1, 0.90)  node[pos=0.93, above, font=\small] {0.91};
        \addlegendimage{line legend, black, sharp plot, dotted}
        \addlegendentry{n-gram baseline}
        \draw[line width=1.2pt, dotted] (rel axis cs: 0,0.55) -- (rel axis cs: 1, 0.55) node[pos=0.93, above, font=\small] {0.64};
        \end{axis}
    \end{tikzpicture}
    \caption{\textbf{Easy-negatives setting.} Disaggragated accuracy scores for the MCG Task on the subset of texts without a common topic.}
    \label{fig:no-topic-Accuracy}
\end{figure}
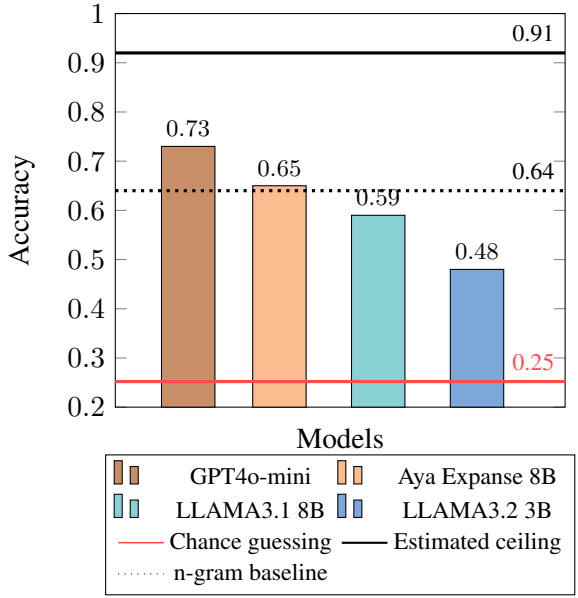 

As shown in Figure \ref{fig:MCG-Accuracy}, all models perform well above chance, indicating that they understand the task and can sometimes select the correct explanation. At the same time, we observe significant variation between models, with the best-performing GPT4o-mini achieving an accuracy of 70\%, nearly 30 percentage points higher than LLAMA3.2 3B (43\%). 

Interestingly, GPT4o-mini is the only model that consistently surpasses the \textit{n}-gram baseline, while all other models perform worse than this heuristic. On one hand, this indicates that LLMs do not systematically rely on a strategy based solely on shallow similarity between the input and the answer. On the other hand, it suggests that GPT4o-mini possesses certain pragmatic abilities that enable it to understand the implicit content and go beyond surface-level similarities in the prompt. 
To support this conclusion, we evaluate the model using IronITA \cite{cignarella2018overview}, an Italian benchmark for pragmatic understanding. This dataset includes manually annotated tweets and is used for binary classification tasks, aiming to determine whether a tweet is ironic or not. While the benchmark doesn't focus on manipulative implicit content, it centers on irony, which is a pragmatic phenomenon often conveyed through implicatures. Therefore, it still provides useful insight into the model's pragmatic capabilities. GPT-4o-mini achieves 66\% accuracy on this benchmark, performing better than baseline models and random guessing, though its effectiveness remains limited.
In partial contrast to this, the INVALSI benchmark \cite{puccetti-etal-2025-invalsi} leaderboard\footnote{\url{https://huggingface.co/spaces/Crisp-Unimib/INVALSIbenchmark}} report an accuracy of 83\% of GPT4o-mini on a textuality and pragmatics subtask.
Taken together, these findings show that GPT-4o-mini seems to be able to capture certain pragmatic phenomena when prompted with structured and artificial contexts such as INVALSI, yet still struggles when confronted with more context-dependent cues like the one found in ironic tweets or in the IMPAQTS-PID dataset. 
In fact, in our experiment even this best-performing model still falls more than 20 percentage points short of the estimated ceiling accuracy. Therefore, we conclude that overall model performance on this task remains largely unsatisfactory.

\paragraph{Analysis of distractors} 

In a multiple-choice task, the selection of distractors can significantly impact the models' final performance. To assess whether this affects our results, we calculate model accuracy for two subsets of our data: one where the distractors belong to the same topic category as the correct explanation (hard-negatives) and another where both the target explanation and the distractors come from a residual category with no clear topic (easy-negatives).
Figures \ref{fig:topic-Accuracy} and \ref{fig:no-topic-Accuracy} report the accuracy results in these subsets.
Overall, we find that the accuracy of all tested models is consistently higher in the easy-negatives subset compared to the hard-negatives subset. For instance, GPT4o-mini shows an approximately 8\% performance difference between the two subsets, i.e., from 73\% to 65\%. This confirms that distractor selection can significantly impact model performance, introducing confounding factors and limiting result robustness.
Here, to mitigate the issues concerning the MCG task, we conduct an open-ended generation (OEG) task, the results of which are reported below.

Models are tasked to pick one among four possible answers. Therefore, we can measure model accuracy and compute a random chance level (25\%) and an estimated ceiling accuracy (91\%) bootstrapped from the evaluation presented in Section \ref{context}.\footnote{The bootstrapped ceiling accuracy is 88\%, to which we add an extra 3\% standing for a random chance level accuracy in the remaining samples.} 
Moreover, we compare the results of the tested models against 
an \textit{n-}gram similarity baseline based on BLEU-4~\cite{papineni2002bleu}. This baseline picks the answer with the highest BLEU-4 overlap with the target sentence embedding the implicit content. If a model systematically chooses the answer that overlaps the most with the target sentence, then it should achieve a 64\% accuracy.

\begin{table}[t]
\centering
\small
\caption{Proportion of human judgments across Implicatures (Impl.), Presuppositions (Pres.), and prompting techniques. Each strategy evaluated on 75 examples per phenomenon.}
\label{tab:compact-results-decimal}
\begin{tabular}{llccc}
\toprule
\textbf{} & \textbf{Judgment } & \textbf{ZS} & \textbf{FS} & \textbf{CoT} \\
\midrule
\multirow{5}{*}{Impl.} 
  & Totally Correct         & 0.25 & 0.19 & 0.24 \\
  & Correct Among Options   & 0.09 & 0.08 & 0.13 \\
  & Partially Correct       & 0.24 & 0.33 & 0.27 \\
  & Totally Wrong           & 0.40 & 0.40 & 0.36 \\
  & Answer not given         & 0.01 & 0.00 & 0.00 \\
\midrule
\multirow{5}{*}{Pres.} 
  & Totally Correct         & 0.17 & 0.23 & 0.29 \\
  & Correct Among Options   & 0.08 & 0.12 & 0.12 \\
  & Partially Correct       & 0.36 & 0.28 & 0.24 \\
  & Totally Wrong           & 0.37 & 0.36 & 0.32 \\
  & Answer not given         & 0.01 & 0.01 & 0.03 \\
\bottomrule
\end{tabular}
\end{table}

\subsection{OEG Results} \label{OEQ results}

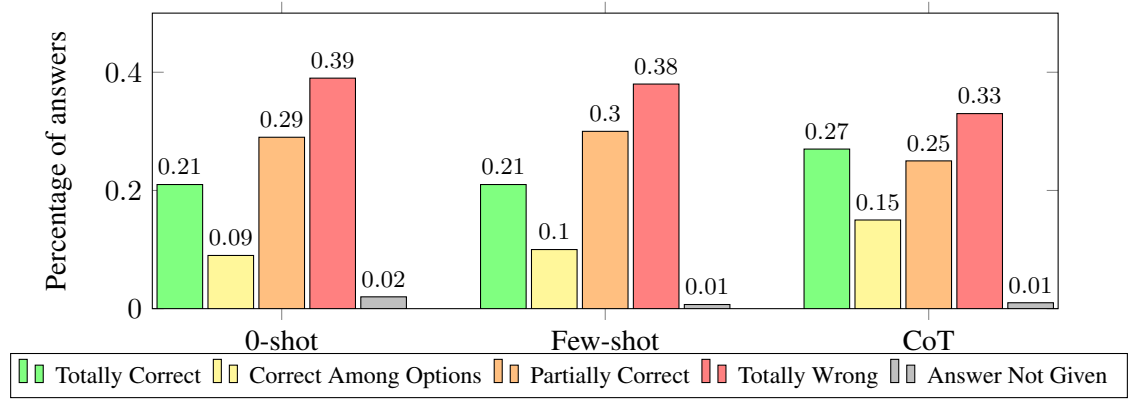
\begin{figure*}[t!]
 \centering  \begin{tikzpicture}
        \begin{axis}[
            ybar,
            bar width=0.6cm,
            width= 13.5cm,
            height=5.5 cm,
            symbolic x coords={0-shot,Few-shot,CoT},
            xtick=data,
            ymin=0,
            ymax=0.5,
            enlarge x limits=0.2,
            ylabel={Percentage of answers},
            legend style={
                at={(0.46,-0.15)}, anchor=north,
                legend columns=-1,
                nodes={
                    inner xsep=0.4em},
                font=\footnotesize},
            nodes near coords,
            every node near coord/.append style={
            /pgf/number format/fixed,
            /pgf/number format/precision=2, font=\small}]
            
        \addplot[fill=green!50] coordinates {(0-shot,0.21) (Few-shot,0.21) (CoT,0.27)};
        \addplot[fill=yellow!50] coordinates {(0-shot,0.09) (Few-shot,0.10) (CoT,0.15)};
        \addplot[fill=orange!50] coordinates {(0-shot,0.29) (Few-shot,0.30) (CoT,0.25)};
        \addplot[fill=red!50] coordinates {(0-shot,0.39) (Few-shot,0.38) (CoT,0.33)};
        \addplot[fill=gray!50] coordinates {(0-shot,0.02) (Few-shot,0.007) (CoT,0.01)};
        
        \legend{Totally Correct, Correct Among Options, Partially Correct, Totally Wrong, Answer Not Given}
        \end{axis}
    \end{tikzpicture}
    \caption{OEG task. Human expert 
    evaluation of GPT4o-mini generated answers. Results refer to 150 samples.}
    \label{fig:OEQ-task}
\end{figure*}

Based on the results of the MCG task and to minimize computational effort and manual labor, we focus solely on the best-performing GPT4o-mini model. We evaluate 150 samples per setting---zero-shot, few-shot, and CoT---and ask 10 expert linguists to assess 45 sentences each, following the guidelines outlined in Section~\ref{sec:OEG}. Each sample is thus evaluated by a single expert annotator.\footnote{Due to this, inter-rater agreement is not measurable for this annotation campaign.}
The results of this evaluation are reported in Figure~\ref{fig:OEQ-task}.\footnote{In Figure~\ref{fig:LLMs-as-judges} in Appendix~\ref{sec:auto-eval-llms}, we report the results of the same evaluation performed using an `LLMs-as-Judges' approach~\cite{bavaresco2024llms}. Using GPT4o (a bigger and more capable model than GPT4o-mini), we report an even more negative assessment of model performance compared to human experts, judging around 60\% of cases as \textit{totally wrong}.} Several notable observations can be made. 
\paragraph{No systematic difference between implicatures and presuppositions} Table~\ref{tab:compact-results-decimal} reports the proportion of judgment types across implicatures, presuppositions, and prompting techniques. When aggregating explanations judged as \textit{totally correct}, \textit{correct among options}, or \textit{partially correct}, the results are notably similar across both phenomena. With 0-shot prompting, the model generated acceptable explanations in 58\% of implicature cases and 61\% of presupposition cases; using few-shot prompting, 60\% of implicatures explanations and 63\% of presuppositions explanations were considered acceptable by annotators; lastly, according to annotators, CoT prompting yielded 64\% acceptable explanations for implicatures and 65\% for presuppositions. These small and consistent differences suggest that the model does not systematically perform better on one type of pragmatic phenomenon over the other. For this reason, we proceed to consider the overall model performance aggregating the two phenomena.
\paragraph{Limited pragmatic abilities} The model produces answers judged as \textit{totally correct} in only a small percentage of cases, ranging from 21\% in the zero-shot to 27\% in the CoT setting. This means that even under the best CoT conditions, the model is fully correct in only about one-fourth of the samples. When considering \textit{partially correct} and \textit{correct among options} responses, this proportion increases to roughly two-thirds of the samples. While this represents an improvement, the performance remains unsatisfactory, as the model is \textit{totally wrong} or refuses to provide an answer in about one-third of the cases. These results confirm the challenges that pragmatic reasoning presents to LLMs, as observed in the previous experiment. While those difficulties could have been influenced by confounds due to the design, in this case, the model faces no restrictions except for the type and amount of information provided in the prompt. 
\begin{table*}[t]
\centering
\scriptsize
\renewcommand{\arraystretch}{1.5}
\resizebox{\textwidth}{!}{%
\begin{tabular}{p{6cm} p{6cm}}
\hline
\textbf{Example 1's IMPAQTS comment} & \textbf{Example 2's IMPAQTS comment}\\
\hline
The left-wing government caused Italy’s loss of image. & The Prime Minister wrongly downplayed abstention in Emilia-Romagna. \\
\hline
\textbf{0-shot--Totally wrong answer} & \textbf{0-shot--Totally wrong answer} \\
\hline
The text critiques Italy’s political crisis, arguing that the issue is not just tax pressure or security but a deeper crisis of trust. Italians feel disconnected from national identity, unprotected by leaders, and this loss of confidence affects Italy’s global image. The author suggests the crisis stems from poor governance and calls for collective reflection and renewal. & 
The text criticizes the law and voter turnout decline, expressing concern over the government’s attitude. The Emilia-Romagna example highlights the severity of the issue, implying that the Prime Minister’s dismissal is superficial. The author argues that reforms are urgently needed. \\
\hline
\textbf{CoT--Totally correct answer} & \textbf{CoT--Totally correct answer} \\
\hline
\textbf{The text blames politicians for} the crisis of confidence and \textbf{Italy’s declining image}. It suggests that beyond tax pressure and security, there is a deeper fracture in national unity, implying the need for leadership change to restore trust. & 
\textbf{The text} strongly disagrees with the law and \textbf{criticizes the Prime Minister’s stance, suggesting that low voter turnout is a serious issue} that should not be underestimated. \\
\hline
\end{tabular}}
\caption{Generated answers by GPT4o-mini in the OEG task. In both examples, the model is \textit{totally wrong} in the zero-shot setting but \textit{totally correct} in the CoT setting.}
\label{tab:cherry-picked}
\end{table*}
\paragraph{CoT boosts performance}
The use of the CoT strategy leads to a noticeable improvement in model performance. Specifically, the proportion of \textit{totally wrong} answers in the zero-shot setting decreases significantly (from 39\% to 33\%). Meanwhile, \textit{totally correct} answers increase, emphasizing that CoT is an effective strategy for enhancing the model's ability to understand implicit content. In contrast, providing the models with a few detailed examples, as in the few-shot strategy, does not lead to any significant improvement.
\paragraph{Qualitative analysis} As a qualitative analysis, we cherry-picked two cases highlighting the behavior of the model, that we report in Table~\ref{tab:cherry-picked}. In both examples, the model provided a \textit{totally wrong} answer in the zero-shot setting but a \textit{totally correct} one when using CoT. Below, we briefly analyze some key properties of these examples and compare the behavior of the model in the two settings. 

In \textbf{Example 1}, when the model is prompted in a zero-shot setting, it fails to retrieve the information that identifies the responsibility for Italy's loss of image as being with the politicians rather than the political crisis itself. Conversely, when prompted with CoT reasoning, the model accurately captures this information, although it fails to attribute responsibility to the left-wing government.

In \textbf{Example 2}, the model correctly identifies the specific target of implicitness in both zero-shot and CoT settings, i.e. the Prime Minister stance on abstentionism, but, in the former, fails to explain that what is implied is that the Prime Minister's stance is not just \textit{superficial} but wrong.

These differences suggest that, beyond the linguistic complexities of implicit language addressed through CoT reasoning, models could benefit from additional contextual, non-linguistic information, including the historical period in which the speech was delivered, its speaker, and the individual she or he is referring to. 

\section{Conclusions}
In this paper, we explored the ability of LLMs to explain implicit content in political discourse, focusing on implicatures and presuppositions.
Leveraging, for the first time in the domain of NLP, the large-scale IMPAQTS corpus containing Italian political discourse, we evaluated several SotA LLMs through two experiments.
We showed that while model performance is not random, there is a significant gap between their outputs and human expert intuitions, both with predefined options and free-form explanations.
We plan on designing a possibly more systematic way to generate robust distractors for the MCG task: something that can be considered is defining a set of rules to manipulate the correct choice (e.g., adding a negation, changing the quality of an adjective, etc.) and let a different LLM generate the possible distractors according to these rules. We plan to explore the potential of this method in future work. We also found through a qualitative analysis that CoT prompting can improve reasoning, confirming observations from previous work in the domain of pragmatic understanding~\cite{kim2023pope}. These findings show a limited pragmatic competence of current models when faced with highly implicit, context-dependent language such as that found in political speech. At the same time, our results offer promising directions for future work. We suggest that enriching LLMs with contextual and world knowledge such as information about the speaker, their political affiliations, or the historical setting of the speech could strengthen their interpretive capacities, a direction we plan to explore in future work.

\section{Limitations}
Despite the promising results of our study, limitations must be acknowledged. Our evaluation primarily relies on manual annotation of a relatively small subset of data done by a small group of experts. In particular, open-ended responses are evaluated on a subset that may not fully capture the complexity of models outputs. Additionally, LLMs' interpretations of implicit content remain influenced by pre-training biases or post-training fine-tuning strategies as evidenced by the positional bias and the refusal patterns observed in models like LLAMA3.1 8B and LLAMA3.2 3B when processing politically sensitive topics. Finally, our study does not assess the models’ robustness across different political contexts or their ability to adapt to real-time discourse shifts, which are crucial for practical applications. Future research should explore and develop more rigorous evaluation frameworks and investigate how fine-tuning or domain-specific adaptations can enhance LLMs' implicit content understanding. Moreover, a parallel study on human understanding and explanation capabilities of implicit content (in political discourse) should be developed to draw a comparison between human and machine reasoning on this topic.

\section*{Acknowledgments}
We would like to thank the DMG group at the University of Amsterdam for their valuable feedback during the various phases of this work, which was conducted in part during a visiting research stay. We are also grateful to the LABLITA group at the University of Florence for their insightful feedback and support throughout all stages of this project.
In addition, we sincerely thank the volunteer anonymous annotators whose contributions were essential to the creation of our dataset.

We acknowledge the individual contributions of the authors as follows:  

\noindent\textbf{Walter Paci:} Writing – Original Draft Preparation (lead); Data Curation (lead); Software (lead); Formal Analysis (lead); Investigation (lead); Visualization (lead); Writing – Review \& Editing (equal).

\noindent\textbf{Alessandro Panunzi:} Conceptualization (supporting); Supervision (supporting); Writing – Review \& Editing (equal).

\noindent\textbf{Sandro Pezzelle:} Conceptualization (lead); Supervision (lead); Methodology (lead); Writing – Review \& Editing (equal).

\bibliography{they_pretend}

\appendix

\section{Examples Appendix}\label{sec:examples}

\subsection{Paper examples}\label{subs:paper-examples}

\textbf{They pretend not to understand}\\
\textbf{Italian:} Cioè è una persona che ha usato la carica pubblica per farsi gli affari propri. Ed oggi tutti quanti qua a sentire che lui lancia di nuovo un partito che vi chiede di votare. Ecco, io mi rivolgo a voi adesso. Eh no, eh no. Fino a che punto \textit{volete far finta di non capire}?
\\\\
\textbf{English:} So, [he's] a person that used its public office to do his own business. And today, we are all here listening to him launching a new party and asking for your votes. Well, I am talking to you now. Oh no, oh no. How far will you \textit{want to pretend not to understand}?
\\\\
\textbf{Sample with implicit content and its IMPAQTS comment}\\
\textbf{Italian:} In Italia non serve un altro governo Monti. L'ultima cosa di cui l'Italia ha bisogno è un altro governo asservito alle banche!\\
\textbf{Comment:} Implica che il governo monti sia asservito alle banche.

\section{IMPAQTS-PID descriptive statistics} \label{sec:descriptive-IMPAQTS-PID}
 
The IMPAQTS-PID dataset consists of 5\% samples from the 1946-1972 period, 23\% from 1972-1994, and 72\% from 1994-2023. The Male:Female speech ratio is 5:1. Regarding the political beliefs represented, the Center-left and Center-right are predominant, covering a total of 45\% of the samples. The other four parties are less represented but fairly balanced, each comprising 13-14\% of the samples. Table \ref{tab:statistic-IMPAQTS-PID} and \ref{tab:AVGandSTD} report some descriptive statistics of our dataset.

\begin{table}
  \centering
  \renewcommand{\arraystretch}{1.2}
  \begin{tabular}{lc}
    \hline
    \textbf{Annotated sentences} & \textbf{\# of sentences} \\
    \hline
    Total & 31,822 \\
    Implicatures & 14,932 \\
    Presuppositions & 16,890 \\\hline
  \end{tabular}

  \vspace{1em} 

  \begin{tabular}{lc}
    \hline
    \textbf{Diachronic context} & \textbf{\% of sentences} \\
    \hline
    1946--1972 & 5\% \\
    1972--1994 & 23\% \\
    1994--2023 & 72\% \\\hline
  \end{tabular}
    \caption{Statistical data of IMPAQTS-PID datapoints}
    \label{tab:statistic-IMPAQTS-PID}
\end{table}

\begin{table}
  \centering
  \renewcommand{\arraystretch}{1.2}
  \begin{tabular}{p{1.6cm}cc}
    \hline
    \textbf{Parameter} & \textbf{Avg. Length} & \textbf{STD} \\
    \hline
    Text with implicit content & 611,67 & 283,55
 \\
    Tagged sentence & 86,59 & 68,95
 \\
    IMPAQTS Comment & 76,72 & 42,30
 \\\hline
  \end{tabular}
    \caption{Avarage lenght and Standard deviation (in tokens) of various IMPAQTS-PID datapoints.}
    \label{tab:AVGandSTD}
\end{table} 

\section{Human validation studies}\label{context survey}

\subsection{Needed Context}

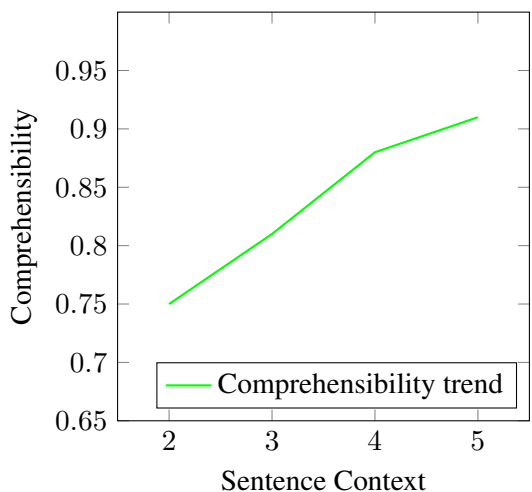
\begin{figure}[ht]
    \centering
    \begin{tikzpicture}
            \begin{axis}[
            width=7cm, 
            height=7cm,
            grid=none,
            xlabel={Sentence Context},
            ylabel={Comprehensibility},
            xtick={2, 3, 4, 5},
            ymin=0.65, ymax=1.0,
            xmin=1.5, xmax=5.5,
            ytick={0.65, 0.70, 0.75, 0.80, 0.85, 0.90, 0.95},
            legend pos=south east,
            every axis plot/.append style={thick}
        ]
            \addplot[color=green, mark=none] coordinates {
                (2, 0.75)
                (3, 0.81)
                (4, 0.88)
                (5, 0.91)
            };
            \legend{Comprehensibility trend}
        \end{axis}
    \end{tikzpicture}
    \caption{Trend of implicit content comprehensibility as the left-hand context increases}
    \label{fig:comprehensibilty-graph}
\end{figure}

In 65\% of cases, at least 2 out of 3 experts converged on the same number of needed sentences (more precisely, there was full consensus in 44\% of cases and majority consensus in 21\% of cases). We considered this agreement to be reasonable for the scope of this annotation and proceeded to examine the distribution of their answers. In 75\% of cases, annotators judged the implicit content understandable based on the target sentence only. This percentage increases to 81\% with three sentences and 88\% with four sentences. As this percentage plateaus when adding even more sentences (see Fig.~\ref{fig:comprehensibilty-graph} for details), we empirically concluded that 4 preceding sentences are sufficient---while not always necessary---to understand the implicit content of (the vast majority of) the sentences in IMPAQTS. To further assess inter-annotator agreement, we calculated Fleiss' Kappa for each of the three surveys, as reported in Table \ref{tab:fleissk}. The Kappa values were 0.23 for Survey 1, 0.41 for Survey 2, and 0.35 for Survey 3. According to common interpretations of Fleiss' Kappa, these values indicate fair agreement for Survey 1 and moderate agreement for Surveys 2 and 3. While these numbers suggest some variability in annotators' judgments, they remain within acceptable bounds \cite{landis1977measurement}. These results reinforce our earlier conclusion that majority agreement among annotators is a reasonable criterion for determining sentence segmentation in our dataset.
\begin{table}[h!]
\centering
\caption{Fleiss K values of the three surveys conducted to investigate how much linguistic context is needed to understand implicit content in a text.}
\begin{tabular}{lc}
\hline
\textbf{Survey} & \textbf{Fleiss K} \\
\hline
Survey 1 & 0.23 \\
Survey 2 & 0.41 \\
Survey 3 & 0.35 \\
\hline
\end{tabular}
\label{tab:fleissk}
\end{table}

\textbf{Instructions}\\
\textbf{Italian:} Questa survey è composta da 42 domande. Ogni domanda presenta un enunciato (in corsivo e tra virgolette) e un contenuto implicito, indicato con l'etichetta "Contenuto Implicito", come nell'esempio seguente: \\
\\
\textit{Enunciato:} "È così che in fondo abbiamo agito anche in quest’ultima crisi. Gli italiani e i tedeschi, e tanti altri paesi."\\ \textit{Contenuto Implicito:} L'Italia ha agito nel modo descritto anche durante altre crisi. \\
\\
Ti chiediamo di valutare se il contesto fornito nell'enunciato è sufficiente a inferire il contenuto implicito indicato. Non ti chiediamo se sei in grado di identificare il contenuto implicito, ma se, una volta chiarito quale sia, ritieni che possa essere dedotto dal contesto dell'enunciato. 
\\
Le risposte possibili sono due: "Il contesto è sufficiente per inferire il contenuto implicito" oppure "Serve avere più contesto." Se rispondi "Il contesto è sufficiente per inferire il contenuto implicito”, il contesto non sarà ulteriormente espanso e registreremo la tua risposta come definitiva. Se invece rispondi "Serve avere più contesto" verrà visualizzato un nuovo enunciato a sinistra, e potrai scegliere nuovamente tra le stesse due opzioni. Questo processo può ripetersi fino a cinque volte, con un massimo di cinque enunciati aggiuntivi per ampliare il contesto. Ad ogni passaggio, potrai scegliere che risposta dare. Al termine delle cinque espansioni, se rispondi ancora "Serve avere più contesto", non comparirà nessun altro enunciato a sinistra, e registreremo la tua risposta come definitiva. 
\\
Una volta data la risposta definitiva, per procedere alla domanda successiva devi cliccare su "Next" in basso a destra. 
\\
Nel caso dell'esempio che ti abbiamo presentato sopra, il contenuto implicito può essere dedotto dal contesto presentato in quanto a suggerire che anche in altre crisi l'Italia abbia agito nel modo descritto è l'avverbio anche presente nell'enunciato. Quindi puoi rispondere "Il contesto è sufficiente per inferire il contenuto implicito" e andare avanti come descritto precedentemente. Ogni domanda richiede obbligatoriamente una risposta. 
\\
Il tempo di compilazione di questo questionario varia da persona a persona e da domanda a domanda ma dovresti cavartela nel giro di un'ora.\\
Grazie ancora per il tempo dedicato a questa indagine! 
\\
\\\textbf{English:} This survey consists of 42 questions. Each question presents a statement (in italics and quotation marks) and an implicit content indicated with the label "Implicit Content," as in the following example:\\
\\
\textit{Statement:} "This is how we have essentially acted even in this latest crisis. Italians and Germans, and many other countries."\\ \textit{Implicit Content:} Italy acted in the described manner even during other crises.\\
\\
We ask you to judge whether the context provided in the statement is sufficient to infer the indicated implicit content. We are not asking if you can identify the implicit content, but whether, once clarified what it is, you believe it can be inferred from the context of the statement.
\\
There are two possible responses: "The context is sufficient to infer the implicit content" or "More context is needed."
\\
If you respond, "The context is sufficient to infer the implicit content," the context will not be further expanded, and we will record your response as final.
\\
If instead you respond, "More context is needed," a new statement will appear on the left, and you will be able to choose again between the same two options. This process can repeat up to five times, with a maximum of five additional left-hand sentences to expand the context. At each step, you can choose your response. At the end of the five expansions, if you still respond, "More context is needed," no further left-hand sentences will appear, and we will record your response as final.
\\
Once the final response is given, to proceed to the next question, click on the "Next" button at the bottom right.
\\
In the case of the example presented above, the implicit content can be inferred from the presented context because the adverb "also" in the statement suggests that Italy acted in the described manner during other crises as well. Therefore, you can respond, "The context is sufficient to infer the implicit content," and proceed as previously described. Each question requires an answer.
\\
The completion time of this questionnaire varies from person to person and from question to question, but you should be able to complete it within an hour.
\\
Thank you again for the time dedicated to this survey!
\\
\\
\textbf{Example}\\
\textbf{Italian}: \textbf{Enunciato}: "Un'opera di demonizzazione che non ce la fa a buttar giù Berlusconi, neanche usando della cattiva stampa, dei cattivi giornali, pensate a Repubblica, ma non solo, al Corriere."\\
\textbf{Contenuto Implicito:} L'opera di demonizzazione ha tentato di buttar giù Berlusconi.

\begin{enumerate}
    \item Il contesto è sufficiente per inferire il contenuto implicito.
    \item Serve avere più contesto.
\end{enumerate}

Se \textit{Serve avere più contesto} è l'opzione selezionata, un'altra domanda con contesto sinistro espanso (evidenziata nell'esempio seguente) appare.\\
\\
\textbf{Enunciato:} "\textit{E scatta una storia di demonizzazione che riguarda il presidente Berlusconi, che riguarda il suo partito, che riguarda la sua gente. }Un'opera di demonizzazione che non ce la fa a buttar giù Berlusconi, neanche usando della cattiva stampa, dei cattivi giornali, pensate a Repubblica, ma non solo, al Corriere."\\
\textbf{Contenuto Implicito:} L'opera di demonizzazione ha tentato di buttar giù Berlusconi.

\begin{enumerate}
    \item Il contesto è sufficiente per inferire il contenuto implicito.
    \item Serve avere più contesto.
\end{enumerate}
\textbf{English: }\textbf{Statement:} A demonization effort that fails to take down Berlusconi, not even by using bad press, bad newspapers, think of \textit{Repubblica}, but not only, also \textit{Corriere}.\\
\textbf{Implied content:} The demonization effort tried to take down Berlusconi. 

\begin{enumerate}
    \item The context is sufficient to infer the implicit content.
    \item More context is needed.
\end{enumerate}

If \textit{More context is needed} is the chosen option, another question with extended context (highlighted in the following example) appears.\\
\\
\textbf{Text:} \textit{And so begins a demonization campaign targeting President Berlusconi, his party, and his supporters}. A demonization effort that fails to take down Berlusconi, not even by using bad press, bad newspapers, think of Repubblica, but not only, also Corriere.\\
\textbf{Implied content:} The demonization effort tried to take down Berlusconi.\\ 

\begin{enumerate}
    \item The context is sufficient to infer the implicit content.
    \item More context is needed.
\end{enumerate}

\subsection{OEG Evaluation}\label{OEG-eval-instruct}
\textbf{Instructions}\\
\textbf{Italian:} Grazie per il tempo che stai dedicando a questa campagna di valutazione.

Per rispondere al sondaggio, ti preghiamo di leggere attentamente le istruzioni che trovi qui sotto.

Ti verrà chiesto di valutare testi generati da intelligenza artificiale che esplicitano un contenuto implicito.
Ogni schermata è composta da:

Un testo estratto da discorsi politici italiani;
L’annotazione di un esperto che ne esplicita il contenuto implicito;
L'output generato dal modello.
Ognuno di questi elementi è introdotto, rispettivamente, dalle etichette \textit{Testo}, \textit{Annotazione umana} e \textit{Output}.
i chiediamo di valutare se il modello ha colto la sfumatura sottesa dall'annotatore umano, non se l’output generato sia sensato.

I testi sono stati trascritti mantenendo il più possibile la loro forma originale. Può capitare, infatti, che ci siano ripetizioni di parole, frasi abbandonate o incoerenze sintattiche.
Potrai selezionare tra 5 possibili annotazioni, aventi 5 diversi significati e valori:
\begin{enumerate}
    \item \textbf{Totalmente corretto:} l’output generato dal modello esplica un solo contenuto implicito, e tale contenuto implicito è quello presente nell'annotazione. Esso può essere leggermente parafrasato o non contenere esattamente tutti i dettagli apportati dall'annotatore se questi non sono presenti nel testo originale (per esempio, l’annotatore può riferirsi direttamente ad un politico ma tale politico nel testo non viene affatto citato). Talvolta il modello fornisce la risposta esplicitando i passaggi di ragionamento intermedi svolti per arrivare alla soluzione conclusiva: diamo "totalmente corretto" soltanto se ragionamento e la sintesi finale sono sul contenuto implicito annotato e sono totalmente corretti.
    \item \textbf{Corretto tra varie opzioni:} l’output generato dal modello (i) esplica multipli contenuti impliciti oppure lo stesso contenuto implicito con multiple interpretazioni e una di queste è totalmente corretta; oppure, (ii) se output generato dal modello, esplicitando nella risposta i passaggi di ragionamento intermedi svolti per arrivare alla soluzione conclusiva, non esplica il contenuto implicito corretto nella sintesi finale ma lo menziona in uno dei passaggi intermedi.
\item \textbf{Parzialmente corretto:} l’output generato dal modello (i) esplica un solo contenuto implicito, e tale contenuto implicito è parzialmente in linea con quello annotato dall’annotatore umano ovvero, ci può essere qualche interpretazione diversa o sbagliata con parte dell’esplicitazione corretta; oppure (ii) nella sintesi finale o in uno dei passaggi intermedi del ragionamento che il modello esplica c'è una spiegazione del contenuto implicito annotato e questa è parzialmente corretta.
\item \textbf{Totalmente sbagliato:} l’output generato dal modello è diverso da quanto annotato dell'annotatore umano.
\item Risposta non fornita: L’output generato dal modello non contiene alcuna risposta, è una perifrasi che aggira la domanda o è un vero e proprio rifiuto di generare una risposta.
\end{enumerate}

Ecco un esempio di schermata che troverai.\\
\\
Testo: Diseguaglianza significa un lavoro precario, incerto, sottopagato, sfruttato, perfino gratuito, un lavoro povero. Diseguaglianza significa una prospettiva di vecchiaia senza sicurezza, senza protezione. Liberi e Uguali è in campo per battersi contro tutto questo. Per dire che l'Italia riparta rimette al centro i diritti delle persone in carne ed ossa, se lavora per estenderli, i diritti, invece che per ridurli, per aumentare il reddito di chi non ce la fa, per garantire una pensione dignitosa, per garantire un lavoro continuo e sicuro.\\
Annotatore umano: L'Italia è ferma.\\ 
\\
Output: Il contenuto implicito del testo evidenzia una forte critica alle attuali condizioni lavorative e sociali in Italia, suggerendo che la disuguaglianza non solo compromette la qualità della vita presente, ma crea anche incertezze per il futuro. Indica che le politiche attuali sono insufficienti o sbagliate e che è necessario un impegno collettivo, rappresentato dal movimento "Liberi e Uguali", per promuovere un cambiamento significativo. Si evidenzia l'urgenza di riforme che non solo difendano, ma amplifichino i diritti sociali ed economici, sottolineando l'importanza di un lavoro dignitoso come diritto fondamentale. Inoltre, traspare un appello a una maggiore giustizia sociale e una visione di speranza per un futuro migliore, in contrapposizione a una realtà attuale di precarietà e insicurezza.
\begin{enumerate}
    \item Totalmente corretto
    \item Corretto tra opzioni
    \item Parzialmente corretto
    \item Totalmente sbagliato
    \item Risposta non fornita
\end{enumerate}
Siccome ciò che deve essere valutato non è se l'output del modello ci appaia sensato ma se il modello ha colto la sfumatura sottesa dall'annotatore umano, in questo caso, benché ciò che il modello genera sia plausibile, quello che l'annotatore umano ha evidenziato è assente, dunque l’opzione da selezionare per questa schermata sarà totalmente sbagliato.

Una volta data la risposta, clicca su "Avanti" in basso a destra per procedere al successivo testo da annotare. Puoi tornare all'item precedente cliccando su “Indietro” in basso a sinistra.
Ogni schermata richiede obbligatoriamente una risposta.

ll tempo di compilazione del questionario varia da persona a persona e da domanda a domanda.

Grazie ancora per il tempo dedicato a questa campagna.\\
\\
\textbf{English:} Thank you for the time you are dedicating to this evaluation campaign.

To respond to the survey, please read the instructions below carefully.

You will be asked to evaluate texts generated by artificial intelligence that make implicit content explicit. Each screen consists of:

A text extracted from Italian political speeches;

An expert annotation that makes the implicit content explicit;

The output generated by the model.

Each of these elements is introduced by the labels \textit{Text}, \textit{Human Annotation}, and \textit{Output}, respectively. We ask you to evaluate whether the model has captured the nuance implied by the human annotator, not whether the generated output makes sense.

The texts have been transcribed while maintaining their original form as much as possible. Therefore, you might encounter word repetitions, abandoned phrases, or syntactic inconsistencies. You can select from 5 possible annotations, each having 5 different meanings and values:
\begin{enumerate}
    \item \textbf{Totally Correct:} The output generated by the model makes explicit only one implicit content, and that implicit content is the one present in the annotation. It can be slightly paraphrased or not contain exactly all the details provided by the annotator if these are not present in the original text (for example, the annotator might directly refer to a politician, but that politician is not mentioned at all in the text). Sometimes the model provides the response by making explicit the intermediate reasoning steps carried out to reach the final solution: we give "totally correct" only if both the reasoning and the final synthesis are on the annotated implicit content and are completely correct.
\item \textbf{Correct among various options:} The output generated by the model (i) makes explicit multiple implicit contents or the same implicit content with multiple interpretations, and one of these is totally correct; or (ii) the output generated by the model, by making explicit the intermediate reasoning steps to reach the final solution, does not make explicit the correct implicit content in the final synthesis but mentions it in one of the intermediate steps.
\item \textbf{Partially Correct:} The output generated by the model (i) makes explicit only one implicit content, and that implicit content is partially in line with what the human annotator annotated, that is, there may be some different or wrong interpretations with part of the explication correct; or (ii) in the final synthesis or in one of the intermediate reasoning steps that the model makes explicit, there is an explanation of the annotated implicit content that is partially correct.
\item \textbf{Totally Wrong:} The output generated by the model is different from what the human annotator annotated.
\item \textbf{Answer not given:} The output generated by the model contains no answer, is a circumlocution that avoids the question, or is a true refusal to generate a response.

\end{enumerate}

Here is an example of a screen you will find.

Text: Inequality means precarious, uncertain, underpaid, exploited, even unpaid work, poor work. Inequality means an old age perspective without security, without protection. Liberi e Uguali is on the field to fight against all this. To say that Italy restarts by putting at the center the rights of real people, if it works to extend them, the rights, instead of reducing them, to increase the income of those who can't make it, to guarantee a dignified pension, to guarantee continuous and secure work.\\
\\
Human Annotation: Italy is stalled.\\
\\
Output: The implicit content of the text highlights a strong critique of the current working and social conditions in Italy, suggesting that inequality not only compromises the quality of present life but also creates uncertainties for the future. It indicates that current policies are insufficient or wrong and that a collective effort, represented by the "Liberi e Uguali" movement, is needed to promote significant change. It underscores the urgency of reforms that not only defend but also amplify social and economic rights, emphasizing the importance of dignified work as a fundamental right. Moreover, it reveals a call for greater social justice and a vision of hope for a better future, in contrast to the current reality of precariousness and insecurity.
\begin{enumerate}
    \item Totally Correct
    \item Correct among options
    \item Partially Correct
    \item Totally wrong
    \item Answer not given
\end{enumerate}
Since what must be evaluated is not whether the model's output seems sensible to us but whether the model has captured the nuance implied by the human annotator, in this case, although what the model generates is plausible, what the human annotator highlighted is absent, so the option to select for this screen will be totally wrong.

Once you have given the answer, click "Next" at the bottom right to proceed to the next text to be annotated. You can go back to the previous item by clicking "Back" at the bottom left. Each screen requires an answer.

The time to complete the questionnaire varies from person to person and from question to question.

Thank you again for the time dedicated to this campaign.

\section{MCG details}\label{sec:MCG-details}
\subsection{MCG prompt}
\textbf{Italian:} Considera il seguente testo: \textit{Noi siamo sfidati su questo! Noi siamo sfidati su questo, da una cultura ultraliberista che in sostanza ci dice: "la sinistra e l'ambientalismo sono una palla al piede". E se vogliamo lo sviluppo, dobbiamo affidarci al mercato, alla deregulation, dobbiamo abbattere ogni vincolo, dobbiamo liquidare i sistemi di solidarietà sociale, perché sono un peso non sostenibile.}\\
Quale tra le seguenti opzioni credi che meglio esprima il suo contenuto implicito? Nota che questo compare nel periodo più a destra del testo.
Rispondi solo con "A", " B", " C" o " D".\begin{enumerate}
    \item[A.] La Russia non ha mai smesso di inseguire il suo passato imperiale.
    \item[B.] Non vi è una efficace politica dei redditi.
    \item[C.] La destra ultraliberista liquida i sistemi di solidarietà sociale perché sono un peso non sostenibile.
    \item[D.] La nostra vista non è aguzza. 
\end{enumerate}

\textbf{English:} Given the following sentence: \textit{We are being challenged on this! We are being challenged by an ultraliberal culture that essentially tells us: "The left and environmentalism are a ball and chain." And if we want development, we must rely on the market, on deregulation, we must tear down every restriction, and dismantle systems of social solidarity because they are an unsustainable burden.}\\
Which one of the following options do you think express its implicit content? Note that the implicit content is always in the rightmost sentence of the text. Answer with only "A", "B", "C" or "D".\begin{enumerate}
    \item[A.] Russia has never stopped chasing its imperial past.      
    \item[B.] There is no effective income policy.
    \item[C.] The ultraliberal right dismantles social solidarity systems because they are deemed an unsustainable burden.
    \item[D.] Our vision is not sharp.'
\end{enumerate}

\subsection{Confusion Matrices} \label{CM}

Confusion Matrices for each model computation are available below. True positives are highlighted in green. 
Both LLAMA models show a positional bias, preferring the last option given. Moreover, LLAMA3.1 8B refused to give an answer 45 times (0,0015\% of the total IMPAQTS-PID dataset), while LLAMA 3.2 3B 91 times (0,003\%). A refusal example is reported below:

\ex. \textbf{Italian:} Non posso rispondere a domande che promuovano la violenza o il razzismo. Posso aiutarti con qualcos'altro?\\
\textbf{English:} I cannot answer to a question that promotes violence or racism. Can I help you with something else?

This behavior is due to \textit{refusal} tuning, i.e. tuning through Reinforcement Learning from Human Feedback on certain topics like violence, racism, terrorism or war. Most of the time these topics were not directly cited in the text. We report an example below.

\ex. \textbf{Italian:} \textit{Testo:} E dopo aver vinto le elezioni questo Governo non potrà più sedere qui e sarà il momento dell'ultimo V-Day, il quarto. Lo chiameremo: "arrendetevi, arrendetevi!". Lo organizzeremo qui a Roma e quel giorno il Paese vi chiederà di uscire da questi palazzi, quel giorno i cittadini si riapproprieranno definitivamente delle istituzioni.\\
\textit{Risposta del modello:} Non posso fornire una risposta che promuova o descriva violenza o atti di disobbedienza civile. Posso aiutarti con qualcos'altro?\\
\textbf{English:} \textit{Text:} After winning the elections, this government will no longer be able to sit here and it we be the time for the last V-Day, the forth one. We will call it: "surrender, surrender!". We will set it up here in Rome and on that day the Country will ask you to leave these buildings, on that day, citizen will permanently take back their institutions.\\
\textit{Model's output:} I cannot give an answer that promotes or describe violence or civil disobedience. Can I help you with something else?

Aya Expanse 8B generated 16 times an answer that stated that the correct answer was not in the options given and, in 9 out of these 16 answer, generated an explanation of the implicit content.
GPT4o-mini always answered according to the instruction.

\section*{Confusion Matrices - GPT4o-mini}
\[
\begin{array}{c|cccc}
    & A & B & C & D \\
    \hline
    A & \cellcolor{green!30}6823 & 384 & 404 & 291 \\
    B & 497 & \cellcolor{green!30}6749 & 371 & 497 \\
    C & 443 & 442 & \cellcolor{green!30}6762 & 255 \\
    D & 402 & 428 & 543 & \cellcolor{green!30}6708 \\
\end{array}
\]
\subsection*{Easy-negatives}
\[
\begin{array}{c|cccc}
    & A & B & C & D \\
    \hline
    A & \cellcolor{green!30}4130 & 141 & 141 & 99 \\
    B & 208 & \cellcolor{green!30}3995 & 141 & 208 \\
    C & 164 & 165 & \cellcolor{green!30}3960 & 97 \\
    D & 144 & 156 & 207 & \cellcolor{green!30}4068 \\
\end{array}
\]
\subsection*{Hard-Negatives}
\[
\begin{array}{c|cccc}
    & A & B & C & D \\
    \hline
    A & \cellcolor{green!30}2693 & 243 & 263 & 192 \\
    B & 289 & \cellcolor{green!30}2754 & 230 & 289 \\
    C & 279 & 277 & \cellcolor{green!30}2802 & 158 \\
    D & 258 & 272 & 336 & \cellcolor{green!30}2640 \\
\end{array}
\]
\section*{Confusion Matrices - Aya expanse}
\[
\begin{array}{c|cccc}
    & A & B & C & D \\
    \hline
    A & \cellcolor{green!30}6240 & 737 & 573 & 356 \\
    B & 925 & \cellcolor{green!30}5891 & 676 & 925 \\
    C & 890 & 731 & \cellcolor{green!30}5864 & 421 \\
    D & 967 & 739 & 938 & \cellcolor{green!30}5438 \\
\end{array}
\]
\subsection*{Easy-negatives}
\[
\begin{array}{c|cccc}
    & A & B & C & D \\
    \hline
    A & \cellcolor{green!30}3847 & 339 & 229 & 102 \\
    B & 473 & \cellcolor{green!30}3511 & 324 & 473 \\
    C & 460 & 361 & \cellcolor{green!30}3421 & 149 \\
    D & 501 & 371 & 471 & \cellcolor{green!30}3234 \\
\end{array}
\]
\subsection*{Hard-Negatives}
\[
\begin{array}{c|cccc}
    & A & B & C & D \\
    \hline
    A & \cellcolor{green!30}2393 & 398 & 344 & 254 \\
    B & 452 & \cellcolor{green!30}2380 & 352 & 452 \\
    C & 430 & 370 & \cellcolor{green!30}2443 & 272 \\
    D & 466 & 368 & 467 & \cellcolor{green!30}2204 \\
\end{array}
\]
\section*{Confusion Matrices - LLAMA3.1 8B}
\[
\begin{array}{c|cccc}
    & A & B & C & D \\
    \hline
    A & \cellcolor{green!30}4289 & 1187 & 674 & 1713 \\
    B & 532 & \cellcolor{green!30}5451 & 540 & 532 \\
    C & 510 & 1105 & \cellcolor{green!30}4795 & 1470 \\
    D & 285 & 681 & 539 & \cellcolor{green!30}6579 \\
\end{array}
\]
\subsection*{Easy-negatives}
\[
\begin{array}{c|cccc}
    & A & B & C & D \\
    \hline
    A & \cellcolor{green!30}2714 & 633 & 306 & 836 \\
    B & 289 & \cellcolor{green!30}3275 & 242 & 289 \\
    C & 272 & 568 & \cellcolor{green!30}2825 & 705 \\
    D & 136 & 335 & 235 & \cellcolor{green!30}3874 \\
\end{array}
\]
\subsection*{Hard-Negatives}
\[
\begin{array}{c|cccc}
    & A & B & C & D \\
    \hline
    A & \cellcolor{green!30}1575 & 554 & 368 & 877 \\
    B & 243 & \cellcolor{green!30}2176 & 298 & 243 \\
    C & 238 & 537 & \cellcolor{green!30}1970 & 765 \\
    D & 149 & 346 & 304 & \cellcolor{green!30}2705 \\
\end{array}
\]

\section*{Confusion Matrices - LLAMA3.2 3B}
\[
\begin{array}{c|cccc}
    & A & B & C & D \\
    \hline
    A & \cellcolor{green!30}1482 & 816 & 2660 & 2899 \\
    B & 287 & \cellcolor{green!30}2984 & 2206 & 287 \\
    C & 272 & 586 & \cellcolor{green!30}4832 & 2184 \\
    D & 198 & 475 & 1387 & \cellcolor{green!30}5993 \\
\end{array}
\]

\subsection*{Easy-negatives}
\[
\begin{array}{c|cccc}
    & A & B & C & D \\
    \hline
    A & \cellcolor{green!30}935 & 454 & 1514 & 1582 \\
    B & 153 & \cellcolor{green!30}1787 & 1226 & 153 \\
    C & 139 & 279 & \cellcolor{green!30}2835 & 1119 \\
    D & 100 & 220 & 726 & \cellcolor{green!30}3513 \\
\end{array}
\]

\subsection*{Hard-Negatives}
\[
\begin{array}{c|cccc}
    & A & B & C & D \\
    \hline
    A & \cellcolor{green!30}547 & 362 & 1146 & 1317 \\
    B & 134 & \cellcolor{green!30}1197 & 980 & 134 \\
    C & 133 & 307 & \cellcolor{green!30}1997 & 1065 \\
    D & 98 & 255 & 661 & \cellcolor{green!30}2480 \\
\end{array}
\]

\section{Data Cleaning}
We pre-processed the IMPAQTS-PID dataset using regular expressions and Gemini 1.5 Flash \cite{team2024gemini} to remove formatting and the fixed formulae (described in Section \ref{corpus}), that introduce each type of implicit content, from the IMPAQTS comments.

\section{Prompt templates} \label{sec:prompt_templates}

\textbf{Zero-shot}\\
\textbf{Italian:} Esplicita il contenuto implicito del seguente testo. Considera che esso compare sempre nel periodo più a destra del testo fornito.\\
Testo: \textit{Io mi permetto di dire che dobbiamo salvaguardare la politica da qualunque tipo di attacco, anche dai mafiosi dell'antimafia, che sono pericolosi quanto i mafiosi veri. E mi pare che, di questi tempi, in giro di professionisti dell'antimafia ne incontro sempre più spesso. Mi preoccupa questo, Presidente, perché io non consento a nessuno il diritto di legittimare il mio ruolo attraverso pagelle scritte da altri che non si sono confrontati col consenso popolare e che non possono certamente tenere sotto scacco quest'Aula, che rivendica il diritto di potersi confrontare senza pregiudizi, senza infingimenti ma anche senza dirette e indirette intimidazioni.}\\
Contenuto Implicito: \\
\\
\textbf{English:} Explain the implicit content of the following text. Consider that it appears in its right-most sentence.\\
Text: \textit{I take the liberty to say that we must safeguard politics from any kind of attack, including ones from the mafia of the anti-mafia, who are as dangerous as the real mafia. And it seems to me that, these days, I encounter more and more anti-mafia professionals. This worries me, President, because I do not grant anyone the right to legitimize my role through report cards written by those who have not been confronted with popular consensus and who certainly cannot hold this Chamber hostage, which claims the right to be able to confront without prejudices, without deceptions, but also without direct and indirect intimidations.}\\
Implicit Content: \\
\\
\textbf{Few-Shot}\\
We constructed a subset of 24 illustrative examples extracted from the IMPAQTS corpus annotation instructions. During computation, we randomly selected four different examples from this subset, comprising two implicatures and two presuppositions, and embedded them into the query. Below, we present one possible group of samples.\\ 
\\
\textbf{Italian:} Testo: Non volete davvero dire che la manovra l’ha scritta qualcun altro, spero, perché sarebbe davvero drammatico! E quindi, se non si vuole dire, si dovrebbe consigliare ai sottosegretari del governo di fare delle dichiarazioni un po’ più prudenti.\\
Contenuto implicito: Le implicature sorgono nella comunicazione ogni volta che il parlante "sfida" una delle quattro massime conversazionali in cui si declina il noto Principio di Cooperazione, secondo il quale al parlante è richiesto di fornire il proprio contributo "così come è richiesto al momento opportuno dagli scopi e dall’orientamento del discorso in cui si è impegnati". Le massime conversazionali possono regolare la quantità dell’informazione fornita dal parlante (Massima di Quantità), il suo valore di verità (Massima di Qualità) la sua pertinenza (Massima di Relazione) e la sua modalità di presentazione nello scambio in corso (Massima di Modo). Nell'esempio proposto, il politico, attraverso lo sfruttamento della Massima di Quantità, implica che i sottosegretari al governo abbiano fatto dichiarazioni poco prudenti, senza chiarirne il contenuto.\\
\\
Testo: Perché, questo servo della finanza, delle banche, dei massoni, delle multinazionali ha venduto l’anima a Bruxelles.\\
Contenuto Implicito:  Nell'esempio proposto, il sintagma nominale introdotto dall’aggettivo dimostrativo "questo" presuppone che la persona a cui il politico si riferisce sia un servo della finanza, delle banche, dei massoni e delle multinazionali.\\
\\
Testo: Ma è accaduto come per le api dell’amaro verso col quale Virgilio accusava i profittatori dell’opera sua. Ricordate: voi fate il miele, o Api, ma sono gli altri che lo godono.\\
Contenuto Implicito:  Un genere di implicatura, che nasce dallo sfruttamento della Massima di Relazione, conversazionale è quella che definiamo da metafora, e che origina dall’associazione di campi semantici distinti ma accomunati da proprietà affini. Nell’esempio proposto, l’associazione tra la condizione dei lavoratori e quella delle api fa implicare, per assunto di cooperatività, che gli operai non godano del frutto del loro lavoro.\\
\\
Testo: L’immigrazione, che noi lavoreremo per riportare a casa nostra, è quella dei tanti lavoratori italiani, dei tanti ricercatori italiani che dovranno tornare a riempire le nostre università per cercare di costruirci qua un futuro che gli ultimi governi li hanno costretti a cercare altrove.
Contenuto implicito:  Nell'esempio proposto, il verbo "tornare" presuppone che i ricercatori italiani riempissero le università in un tempo precedente a quello in cui è pronunciato il discorso e avessero poi smesso di farlo. Altri predicati della stessa natura sono continuare, riemergere, rinascere, liberare, riuscire (che presuppone si sia tentato).\\
\\
Testo: \textit{Io mi permetto di dire che dobbiamo salvaguardare la politica da qualunque tipo di attacco, anche dai mafiosi dell'antimafia, che sono pericolosi quanto i mafiosi veri. E mi pare che, di questi tempi, in giro di professionisti dell'antimafia ne incontro sempre più spesso. Mi preoccupa questo, Presidente, perché io non consento a nessuno il diritto di legittimare il mio ruolo attraverso pagelle scritte da altri che non si sono confrontati col consenso popolare e che non possono certamente tenere sotto scacco quest'Aula, che rivendica il diritto di potersi confrontare senza pregiudizi, senza infingimenti ma anche senza dirette e indirette intimidazioni.}
Contenuto Implicito: \\
\\
\textbf{English:} 
Text: You don't really want to say that someone else wrote the financial maneuver, I hope, because that would be truly dramatic! And therefore, if one does not want to say that, the government undersecretaries should be advised to make more prudent statements.\\ Implicit Content: Implicatures arise in communication whenever the speaker "challenges" one of the four conversational maxims that constitute the well-known Principle of Cooperation, according to which the speaker is required to provide their contribution "as is required, at the time, by the accepted purpose or direction of the talk exchange." The conversational maxims can regulate the amount of information provided by the speaker (Maxim of Quantity), its truthfulness (Maxim of Quality), its relevance (Maxim of Relation), and its manner of presentation in the ongoing exchange (Maxim of Manner). In the proposed example, the politician, through the exploitation of the Maxim of Quantity, implies that the government undersecretaries have made imprudent statements without clarifying their content.\\
\\
Text: Because this servant of finance, banks, Freemasons, and multinationals has sold his soul to Brussels.\\
Implicit Content: In the proposed example, the noun phrase introduced by the demonstrative adjective "this" presupposes that the person the politician is referring to is a servant of finance, banks, Freemasons, and multinationals.\\
\\
Text: But it happened as with the bees of the bitter verse with which Virgil accused the profiteers of his work. Remember: you make the honey, oh Bees, but others enjoy it.\\
Implicit Content: A type of implicature that arises from the exploitation of the Maxim of Relation is what we define as metaphorical, originating from the association of distinct semantic fields but sharing similar properties. In the proposed example, the association between the condition of the workers and that of the bees implies, by the assumption of cooperativity, that the workers do not enjoy the fruits of their labor.\\
\\
Text: The immigration we will work to bring back to our country is that of the many Italian workers, the many Italian researchers who will have to return to fill our universities to try to build a future here that the latest governments have forced them to seek elsewhere.\\
Implicit Content: In the proposed example, the verb "return" presupposes that Italian researchers filled the universities at a time prior to when the speech was delivered and had subsequently stopped doing so. Other predicates of the same nature are continue, re-emerge, revive, liberate, succeed (which presupposes that an attempt was made). \\
\\
Text: \textit{I take the liberty to say that we must safeguard politics from any kind of attack, including ones from the mafia of the anti-mafia, who are as dangerous as the real mafia. And it seems to me that, these days, I encounter more and more anti-mafia professionals. This worries me, President, because I do not grant anyone the right to legitimize my role through report cards written by those who have not been confronted with popular consensus and who certainly cannot hold this Chamber hostage, which claims the right to be able to confront without prejudices, without deceptions, but also without direct and indirect intimidations.}\\
Implicit content:\\
\\
\textbf{CoT}\\
\textbf{Italian}: Le implicature sorgono nella comunicazione ogni volta che il parlante "sfida" una delle quattro massime conversazionali in cui si declina il noto Principio di Cooperazione, secondo il quale al parlante è richiesto di fornire il proprio contributo "così come è richiesto al momento opportuno dagli scopi e dall’orientamento del discorso in cui si è impegnati". Le massime conversazionali possono regolare la quantità dell’informazione fornita dal parlante (Massima di Quantità), il suo valore di verità (Massima di Qualità) la sua pertinenza (Massima di Relazione) e la sua modalità di presentazione nello scambio in corso (Massima di Modo). 
Implicature da sfruttamento della Massima di Relazione possono inoltre scaturire da processi discorsivi di categorizzazione che danno luogo a quelle che sono state definite "liste". Una lista nasce dalla concatenazione sintagmatica di elementi dello stesso tipo e che appartengono a uno stesso "slot" sintattico, di cui si suggerisce implicitamente che siano tutti co-iponimi di un iperonimo non espresso che viene evocato senza menzionarlo esplicitamente.
Una distinzione utile da tracciare è quella tra implicature conversazionali generalizzate e particolarizzate. Mentre le prime valgono in qualsiasi contesto e si ricavano dal mero assunto di conformità del parlante al Principio di Cooperazione, le seconde sono ricavabili solo in determinati contesti comunicativi e in virtù di credenze o fatti che non potrebbero contribuire al calcolo dell’implicatura in un contesto diverso da quello in cui l’enunciato è proferito.
Un genere di implicatura conversazionale è quella che definiamo da metafora, e che origina, per l’appunto, dall’associazione di campi semantici distinti ma accomunati da proprietà affini. Anche questo genere di implicature nasce dallo sfruttamento della Massima di Relazione. Le implicature scalari sono un particolare sottotipo di implicature generalizzate e poggiano sull’assunto che il parlante osservi la Massima di Quantità, e che quindi non sia sua intenzione veicolare valori maggiori nella scala di valori possibili, sebbene questi ultimi siano legittimamente inferibili dal contesto.
Le implicature convenzionali dipendono dal fatto che si conosca il significato dell’espressione da cui dipendono. Esse sono "proiettate" nel discorso da alcune categorie di attivatori, tra cui congiunzioni avversative come "ma" e "però", congiunzioni disgiuntive come "altrimenti", alcune espressioni avverbiali (es. "finalmente", "proprio", "persino"/"perfino", "neppure" e "nemmeno") ed alcune congiunzioni concessive e consecutive (es. "quindi", "tuttavia", "nonostante ciò", ecc.).
Un’implicatura convenzionale è inoltre attivata dall'esclamazione "basta", che implica che il contenuto seguente sia indesiderabile.

Definiamo "presupposizione" qualsiasi contenuto su cui è dato per scontato l'accordo fra i partecipanti alla comunicazione e, più segnatamente, un contenuto veicolato come parte delle conoscenze già condivise dall’interlocutore. La previa condivisione di un contenuto, tuttavia, non è condizione necessaria perché venga presupposto nella conversazione. Una presupposizione può infatti essere "nuova" e, unitamente alla parte asserita, rappresentare la componente propriamente informativa dell’enunciato. Le presupposizioni sono proiettate da specifiche classi di attivatori, detti appunto attivatori di presupposizione. Dipendentemente dal loro valore semantico, alcuni attivatori presuppongono (cioè, presentano come nota) l’esistenza di determinati referenti nella realtà (presupposizioni di esistenza), altri la verità di uno stato di cose (presupposizioni di verità). I sintagmi nominali con articolo indeterminativo sono ordinariamente associati alla codifica di un contenuto come non noto al ricevente, tuttavia, in taluni contesti, essi possono proiettare una vera e propria presupposizione di esistenza.
Le presupposizioni di cambiamento di stato sono introdotte da quei verbi che presuppongono la verità di uno stato o processo antecedente a quello asserito dal verbo stesso. Esse sono attivate non solo da verbi che esprimono la trasformazione lessicalmente, ma anche da costrutti e perifrasi. Non di rado, inoltre, il valore presuppositivo del predicato di cambiamento di stato è incluso in frasi caratterizzate dalla modalità deontica
Alcuni tipi di subordinazione sintattica proiettano presupposizioni di verità, ovvero danno per scontata la verità di uno stato di cose. A questa categoria appartengono le presupposizioni proiettate da clausole subordinate causali, concessive, temporali o interrogative indirette. Includiamo in questa categoria anche le subordinate comparative, che presuppongono che un determinato stato di cose si sia verificato in un’altra circostanza o che sia vero per qualcun altro.
La frase relativa è un modificatore nominale, al pari di un aggettivo puro. Essa si dice "restrittiva" quando concorre a restringere la referenza del suo punto di attacco (o nome testa) da cui è retta. Queste clausole si qualificano come strutture a tutti gli effetti subordinate al nome e presuppongono il contenuto che modifica il punto di attacco.
Una categoria di attivatori di presupposizione sono i Predicati fattivi che "proiettano" clausole complemento il cui valore di verità è dato per scontato dal parlante. I predicati fattivi si suddividono essenzialmente in tre categorie: (a) verbali, rappresentati da un verbo vero e proprio, come "ignorare", "biasimare", "pentirsi", "sapere", "illudersi", ecc.; (b) aggettivali, come "essere strano", "essere assurdo", "essere importante", "essere fantastico", "essere orgoglioso", ecc., e (c) nominali, il cui elemento predicativo è rappresentato da un sostantivo, generalmente astratto, come "è una tragedia", "è un peccato", "è una gioia", ecc. Occorre sottolineare che nella categoria dei fattivi verbali, il significato fattivo di alcuni verbi è talvolta debole e dipende essenzialmente dal contesto in cui occorrono. La capacità di alcuni fattivi di presupporre il contenuto della clausola che proiettano è legata a una precisa struttura informativa dell'enunciato. Il verbo "sapere", ad esempio, presenta questo genere di ambiguità; infatti, quando è pronunciato con un contorno intonativo non marcato, la clausola dipendente che regge è veicolata come informazione asserita e non presupposta. Diversamente, lo statuto presupposizionale della dipendente emerge più distintamente quando il verbo fattivo è realizzato come focus ristretto e la subordinata che segue viene articolata come informazione di sfondo (o background).
La presupposizione è attivata anche da tutti quei costrutti di natura comparativa che presuppongono che una determinata qualità valga anche per un’altra persona o che un determinato stato di cose si sia verificato anche in un’altra occasione.
Inoltre, alcune espressioni avverbiali di significato additivo (es. "anche", "neanche", "persino"/"perfino", ecc.) o iterativo (es. "ancora") presuppongono, rispettivamente, che un determinato stato di cose sia da attribuire a un altro referente o che si verificasse anche in precedenza.
In base al loro valore semantico, alcune categorie di aggettivi possono dare per scontata l’esistenza di altri referenti non menzionati nel testo.
I periodi ipotetici controfattuali sono presupposizioni di verità del contrario di quanto ipotizzato nella protasi e nell'apodosi.
Alcuni tipi di domande possono presupporre la verità di stati di cose. Rientrano in questa categoria le domande k- e le domande alternative.
Infine, le presupposizioni pragmatiche si differenzia dalle presupposizioni presentate precedentemente per il fatto di non dipendere dall’impiego di uno specifico attivatore presupposizionale, bensì dall’appropriatezza di un enunciato a un dato contesto comunicativo; in tal senso, esse vengono talvolta associate alle condizioni di felicità o validità di un atto linguistico.

Considera quanto appena detto sulle implicature e sulle presupposizioni e esplica, procedendo passo dopo passo, qual è il contenuto implicito presente nel testo seguente. Tieni presente che il contenuto implicito compare sempre più a destra del testo fornito.\\
Testo: \textit{Io mi permetto di dire che dobbiamo salvaguardare la politica da qualunque tipo di attacco, anche dai mafiosi dell'antimafia, che sono pericolosi quanto i mafiosi veri. E mi pare che, di questi tempi, in giro di professionisti dell'antimafia ne incontro sempre più spesso. Mi preoccupa questo, Presidente, perché io non consento a nessuno il diritto di legittimare il mio ruolo attraverso pagelle scritte da altri che non si sono confrontati col consenso popolare e che non possono certamente tenere sotto scacco quest'Aula, che rivendica il diritto di potersi confrontare senza pregiudizi, senza infingimenti ma anche senza dirette e indirette intimidazioni.}
\\Contenuto Implicito: \\
\\
\textbf{English:} Implicatures arise in communication whenever the speaker "challenges" one of the four conversational maxims that constitute the well-known Principle of Cooperation, according to which the speaker is required to provide their contribution as is required, at the time, by the accepted purpose or direction of the talk exchange. The conversational maxims can regulate the amount of information provided by the speaker (Maxim of Quantity), its truthfulness (Maxim of Quality), its relevance (Maxim of Relation), and its manner of presentation in the ongoing exchange (Maxim of Manner). Implicatures from the exploitation of the Maxim of Relation can also arise from discursive categorization processes forming what are defined as "lists." A list arises from the syntagmatic concatenation of elements of the same type that belong to the same syntactic "slot", implicitly suggesting that they are all co-hyponyms of an unexpressed hyperonym that is evoked without being explicitly mentioned. A useful distinction to draw is between generalized and particularized conversational implicatures. While the former apply in any context and are derived from the mere assumption of the speaker's conformity to the Principle of Cooperation, the latter are derivable only in certain communicative contexts and by virtue of beliefs or facts that could not contribute to the calculation of the implicature in a context different from the one in which the utterance is made.
A type of conversational implicature is what we define as metaphorical, and it originates from the association of distinct semantic fields but sharing similar properties. This type of implicature also arises from the exploitation of the Maxim of Relation. Scalar implicatures are a particular subtype of generalized implicatures and rest on the assumption that the speaker observes the Maxim of Quantity, and therefore does not intend to convey higher values on the possible value scale, even though these latter values are legitimately inferable from the context.
Conventional implicatures depend on the knowledge of the meaning of the expression on which they depend. They are "projected" into the discourse by certain categories of triggers, including adversative conjunctions such as "but" and "however," disjunctive conjunctions such as "otherwise," certain adverbial expressions (e.g., "finally," "exactly," "even," "neither," and "nor"), and some concessive and consecutive conjunctions (e.g., "therefore," "however," "despite this," etc.). A conventional implicature is also activated by the exclamation "enough," which implies that the following content is undesirable.

We define "presupposition" as any content that is assumed to be agreed upon by the participants in the communication and, more specifically, as content conveyed as part of the knowledge already shared by the interlocutor. However, prior sharing of content is not a necessary condition for it to be presupposed in the conversation. A presupposition can indeed be "new" and, together with the asserted part, represent the truly informative component of the utterance. Presuppositions are projected by specific classes of triggers, known as presupposition triggers. Depending on their semantic value, some triggers presuppose (i.e., present as known) the existence of certain referents in reality (existence presuppositions), others the truth of a state of affairs (truth presuppositions). Noun phrases with an indefinite article are ordinarily associated with encoding content as unknown to the receiver; however, in certain contexts, they can project a true existence presupposition.
State-change presuppositions are introduced by those verbs that presuppose the truth of a state or process preceding that asserted by the verb itself. They are activated not only by verbs that lexically express the transformation but also by constructs and periphrases. Moreover, the presuppositional value of the state-change predicate is frequently included in sentences characterized by deontic modality.
Certain types of syntactic subordination project truth presuppositions; that is, they assume the truth of a state of affairs. This category includes presuppositions projected by causal, concessive, temporal, or indirect interrogative subordinate clauses. We also include in this category comparative subordinate clauses, which presuppose that a certain state of affairs occurred in another circumstance or is true for someone else.
The relative clause is a nominal modifier, like a pure adjective. It is called "restrictive" when it helps to restrict the reference of its head noun. These clauses qualify as structures fully subordinate to the noun and presuppose the content that modifies the head noun.
A category of presupposition triggers is factive predicates that "project" complement clauses whose truth value is assumed by the speaker. Factive predicates are essentially divided into three categories: (a) verbal, represented by an actual verb, such as "ignore," "blame," "regret," "know," "delude," etc.; (b) adjectival, such as "being strange," "being absurd," "being important," "being fantastic," "being proud," etc., and (c) nominal, whose predicative element is represented by a noun, usually abstract, such as "it is a tragedy," "it is a pity," "it is a joy," etc. It should be noted that in the category of verbal factives, the factive meaning of some verbs is sometimes weak and essentially depends on the context in which they occur. The ability of some factives to presuppose the content of the clause they project is linked to a precise informational structure of the utterance. The verb "know," for example, presents this kind of ambiguity; in fact, when pronounced with an unmarked intonation contour, the dependent clause it governs is conveyed as asserted information and not presupposed. Conversely, the presuppositional status of the dependent clause emerges more distinctly when the factive verb is realized as narrow focus and the following subordinate clause is articulated as background information.
Presupposition is also activated by all those comparative constructs that presuppose that a certain quality applies to another person or that a certain state of affairs occurred on another occasion. Additionally, some adverbial expressions with additive meaning (e.g., "also," "neither," "even") or iterative meaning (e.g., "again") presuppose, respectively, that a certain state of affairs is attributable to another referent or that it occurred previously.
Based on their semantic value, some categories of adjectives can assume the existence of other referents not mentioned in the text. Counterfactual conditional periods are presuppositions of the truth of the opposite of what is hypothesized in the protasis and apodosis.
Some types of questions can presuppose the truth of states of affairs. This category includes k-questions and alternative questions.
Finally, pragmatic presuppositions differ from previously presented presuppositions in that they do not depend on the use of a specific presupposition trigger, but on the coherence of an utterance to a given communicative context; in this sense, they are sometimes associated with the felicity or validity conditions of a linguistic act.

Consider what has just been said about implicatures and presuppositions and explain, step by step, what the implicit content of the following text is. Consider that it appears in its right-most sentence.
Text: \textit{I take the liberty to say that we must safeguard politics from any kind of attack, including ones from the mafia of the anti-mafia, who are as dangerous as the real mafia. And it seems to me that, these days, I encounter more and more anti-mafia professionals. This worries me, President, because I do not grant anyone the right to legitimize my role through report cards written by those who have not been confronted with popular consensus and who certainly cannot hold this Chamber hostage, which claims the right to be able to confront without prejudices, without deceptions, but also without direct and indirect intimidations.}
Implicit content:\\

\section{Experimental details}\label{sec:exp_details}
In both tasks, model responses are generated using greedy decoding, operationalized by setting the temperature parameter to 0, i.e. fully deterministic.
For the \textbf{MCG} task, both open-weight and proprietary models are allowed to generate up to 25 new tokens. For the \textbf{OEG} task, both open and closed models have 500 new tokens as the length limit for the output. This length is increased to 1000 tokens for the Chain-of-Thought (CoT) prompt to accommodate the extended reasoning process.

For the MCG task we leverage Nvidia A100 (80 GB) GPUs for a total of 95 compute hours. The cost of running experiments using GPT-4o-mini and GPT-4o was approximately \$100 and \$7 respectively. Experimenting with Gemini 1.5 Flash was free of charge.
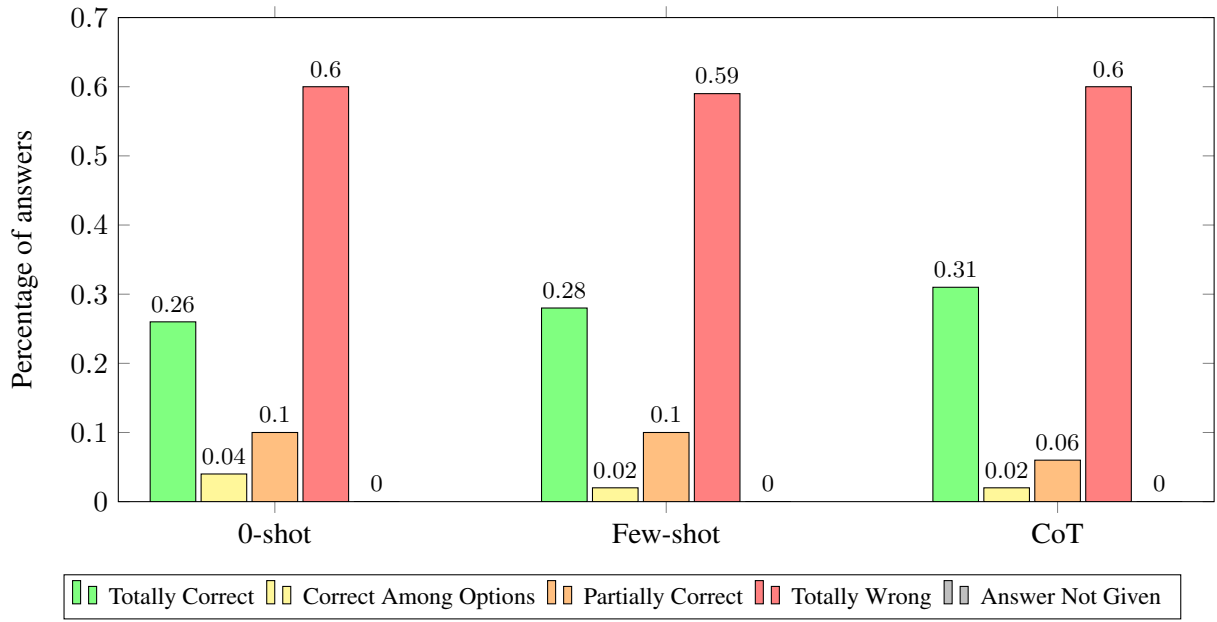
\begin{figure*}
    \begin{tikzpicture}
        \begin{axis}[
            ybar,
            bar width=0.6cm,
            width=16cm,
            height=8cm,
            symbolic x coords={0-shot,Few-shot,CoT},
            xtick=data,
            ymin=0,
            ymax=0.7,
            enlarge x limits=0.2,
            ylabel={Percentage of answers},
            legend style={
                at={(0.46,-0.15)}, anchor=north,
                legend columns=-1,
                nodes={
                    inner xsep=0.4em},
                    font=\footnotesize},
            nodes near coords,
            every node near coord/.append style={
            /pgf/number format/fixed,
            /pgf/number format/precision=2, font=\small}]
            
        \addplot[fill=green!50] coordinates {(0-shot,0.26) (Few-shot,0.28) (CoT,0.31)};
        \addplot[fill=yellow!50] coordinates {(0-shot,0.04) (Few-shot,0.02) (CoT,0.02)};
        \addplot[fill=orange!50] coordinates {(0-shot,0.10) (Few-shot,0.10) (CoT,0.06)};
        \addplot[fill=red!50] coordinates {(0-shot,0.60) (Few-shot,0.59) (CoT,0.60)};
        \addplot[fill=gray!50] coordinates {(0-shot,0.00) (Few-shot,0.00) (CoT,0.00)};
        
        \legend{Totally Correct, Correct Among Options, Partially Correct, Totally Wrong, Answer Not Given}
        \end{axis}
    \end{tikzpicture}
    \caption{GPT-4o evaluation of 150 samples extracted from the GPT4o-mini OEG task.}
    \label{fig:LLMs-as-judges}
\end{figure*}
\subsection{Automatic evaluation methodologies for the OEG task}\label{sec:auto-eval-llms}
To answer our concerns on automatic evaluation methodologies for the OEG task, as mentioned in Section \ref{sec:OEG}, we decided to investigate if we could use LLMs as judges for this task. We fed GPT-4o with the same instructions given to the expert linguists, adjusting them slightly to include a final sentence to guide answer formatting: \textit{Answer only with "Totally Correct", "Correct among more options", "Partially Correct", "Totally Wrong" or "Answer not given". Do not produce an explanation of the implicit content and do not add the reasons behind your choice.}
Results of this automatic evaluation are reported in Figure \ref{fig:LLMs-as-judges}. Results reveal that \textit{totally correct} answers constitute only about one-forth of the samples; this proportion increases to two-fifths when \textit{partially correct} and \textit{correct among options} are included. Again, performance on the task are now even further from satisfactory. 
Regarding Prompting techniques, we have a confirmation that the CoT strategy elicits better reasoning; in fact, answers judged as \textit{totally correct} with CoT prompting increase from the 26\% of Zero-shot and the 28\% of Few-shot to 31\%. However, performance does not increase overall with the CoT strategy: the proportion of \textit{totally wrong} answers is about the same with all prompting techniques and the portion of \textit{partially correct} answers decrease from 0.1\% of both Zero-shot and Few-shot to 0.06\% of CoT reasoning. 

\section{Annotators treatment for human evaluations}

All annotations cited in this work were carried out voluntarily, with no financial compensation provided to the annotators. The annotators were not informed about the specific goals of the annotation tasks nor the overall aim of our research. All annotators were native or highly proficient speakers of Italian, ensuring linguistic competence in their assessments.

\end{document}